\documentclass[10pt,twocolumn,letterpaper]{article}
\usepackage{graphicx}
\usepackage{subcaption}
\usepackage[pagenumbers]{wacv}

\usepackage[subtle]{savetrees} %

\definecolor{wacvblue}{rgb}{0.21,0.49,0.74}
\usepackage[pagebackref,breaklinks,colorlinks,allcolors=wacvblue]{hyperref}
\usepackage{multirow}
\usepackage{pifont}
\newcommand{\tg}{\textcolor{green!60!black}{\checkmark}}
\newcommand{\tr}{\textcolor{red}{\ding{55}}}

\def\wacvPaperID{1759}
\def\confName{WACV}
\def\confYear{2027}

\title{MotoVerse: A Benchmark for Predictive Perception of Motorized Two-Wheelers \\in Dense Unstructured Traffic}

\author{Varun A. Paturkar \qquad Shankar Gangisetty \qquad C.V. Jawahar\\
IIIT Hyderabad, India\\
{\tt\small varuna.paturkar@research.iiit.ac.in, shankar.gangisetty@ihub-data.iiit.ac.in, jawahar@iiit.ac.in}
}

\begin{document}
\maketitle
\begin{abstract}
Motorized Two-Wheelers (MTWs) dominate personal transport across much of the Global South, yet remain largely overlooked in predictive perception research for autonomous driving and ADAS. We introduce MotoVerse, a benchmark for predictive perception from an MTW's ego-centric perspective. MotoVerse comprises 949 maneuver clips collected in dense, unstructured traffic and trajectory annotations for over 31,000 surrounding road agents, including over 16,000 MTWs,
enabling two complementary tasks: Ego-Rider Maneuver Anticipation (ERMA) and Traffic Agent Trajectory Prediction (TATP). Each clip is further annotated with an MTW-specific challenge taxonomy comprising dynamic ego-motion, weak traffic structure and contextual occlusion, enabling challenge-aware evaluation. We benchmark established conventional prediction models, {
general-purpose Vision-Language Models (VLMs), driving-specific VLMs and vision-language-action models}. Our experiments reveal {three} key insights: (i) {video backbones reach most of their full-observation accuracy from the first fifth of a maneuver, whereas zero-shot VLMs need most of the maneuver to be observed}; (ii) {zero-shot VLMs, whether general-purpose or driving-specific, largely remain close to simple baselines on both tasks; a preliminary LoRA study shows that adaptation lifts anticipation sharply but leaves trajectory prediction behind conventional models}; and (iii) {the rider's own motion drives forecasting difficulty: for every model, including a static baseline, errors are highest during left turns and rise under dynamic ego-motion}. We believe MotoVerse will serve as a challenging benchmark for safer and more robust predictive AI for intelligent two-wheelers. The dataset and evaluation code are publicly available at \url{https://github.com/VarunIIITH/MotoVerse}.
\end{abstract}

\section{Introduction}
\label{sec:intro}

Autonomous driving and ADAS have advanced rapidly over the last decade, driven largely by the availability of large-scale, diverse driving datasets~\cite{nuScenes, kitti, bdd100k, waymo_open_dataset}. However, these datasets predominantly capture four-wheeled vehicles operating on well-structured road infrastructure of the Global North. Traffic environments in the Global South present fundamentally different challenges, characterized by unstructured roads, heterogeneous traffic participants, weak lane discipline, and highly dynamic interactions. Although recent datasets such as IDD~\cite{idd} and METEOR~\cite{meteor} have begun to address these complexities, they continue to focus primarily on four-wheelers.

In contrast, \textbf{Motorized Two-Wheelers (MTWs)} are the dominant mode of personal transportation across much of the Global South. The riders are significantly more vulnerable due to minimal physical protection and constant close interactions with other traffic agents. Consequently, MTW users account for a disproportionately large share of road traffic deaths in many countries in the Global South~\cite{morth}. Despite their societal importance and elevated safety risks, the community lacks large-scale datasets and standardized benchmarks tailored to understanding the unique challenges of MTW-centric riding.

\begin{figure*}[t]
    \centering
  \includegraphics[width=\textwidth]{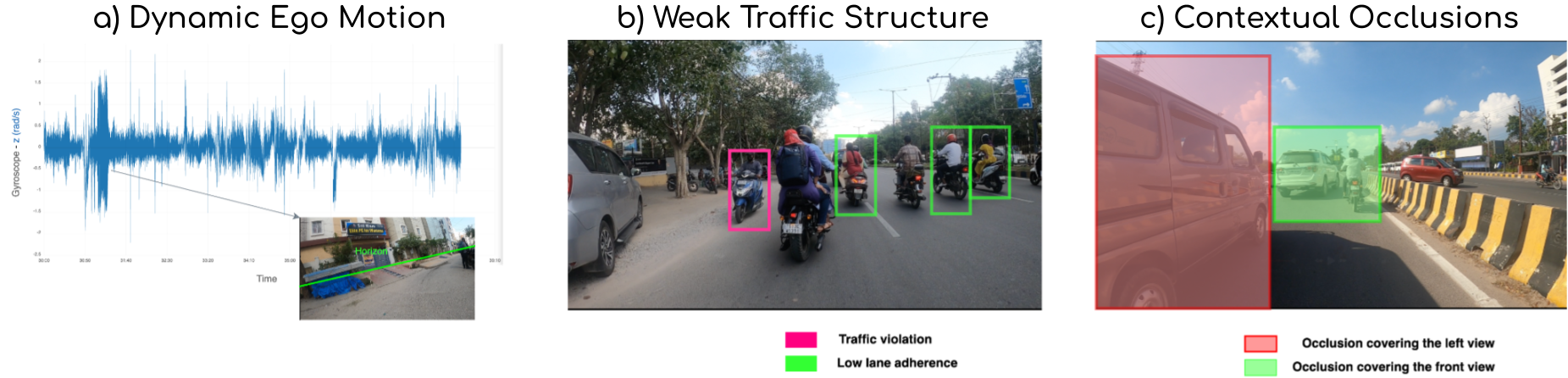}  
   \caption{\textbf{MTW-specific challenges in unstructured traffic.} (a) \textbf{Dynamic ego-motion}: gyroscope measurements illustrating abrupt camera rotations and high-frequency vibrations during rider maneuvers (grey arrow highlights the fluctuations). (b) \textbf{Weak traffic structure}: MTWs frequently deviate from lane structure while navigating dense traffic. (c) \textbf{Contextual occlusion}: close-proximity riding through narrow gaps results in frequent occlusions of surrounding traffic participants.}
    \label{fig:teaser_fig}
\end{figure*}

The sensing available on an MTW also differs fundamentally from that assumed by existing driving datasets. MTWs in the Global South are low-cost, high-volume
vehicles, and the sensor suites used on four-wheelers, including multi-camera rigs, LiDAR, and differential GNSS, cost a substantial fraction of the vehicle and at times more than the MTW itself. Mounting is equally constrained, as a two-wheeler offers no roof or stable chassis surface, exposes any instrumentation to weather and continuous vibration, and competes directly with rider ergonomics. While richer sensor
configurations are feasible for research platforms, a single forward-facing camera remains the only configuration with a realistic path to deployment at scale.

Building safe AI systems for MTWs requires the ability to anticipate both the rider's intent and the behavior of surrounding traffic before critical situations unfold. Accordingly, we focus on two complementary \textbf{predictive perception} tasks: \emph{Ego-Rider Maneuver Anticipation (ERMA)}~\cite{gebert2019end, jain2015car, jain2016recurrent, khairdoost2020real}, which predicts the ego rider's upcoming maneuver from an early portion of the observed video, and \emph{Traffic Agent Trajectory Prediction (TATP)}~\cite{bitrap, ABC+, PIETraj, SGNet}, which forecasts the future
{2D bounding-box} trajectories of
surrounding road agents {in the image plane of} the rider's
ego-centric {camera}.
While recent datasets~\cite{meteor, daad-x} have begun investigating predictive tasks in unstructured traffic, they remain focused on four-wheelers.

Existing prediction methods are developed for four-wheelers with stable camera viewpoints and structured traffic. These assumptions do not hold for MTWs, which face three key challenges (Fig.~\ref{fig:teaser_fig}). \textbf{Dynamic ego-motion:} unlike rigidly mounted
cameras on four-wheelers, MTW-mounted cameras undergo rapid pose and orientation changes due to rider lean, steering corrections and road vibrations, leading to highly non-stationary observations. \textbf{Weak traffic structure:} MTWs
frequently navigate narrow gaps and exhibit limited lane adherence in dense mixed traffic, rendering road geometry and lane-based priors unreliable. \textbf{Contextual occlusion:} operating in close proximity to larger
vehicles results in persistent, dynamic occlusions that obscure critical cues such as surrounding agents, traffic signals and road boundaries. Collectively, these factors distinguish MTW riding from four-wheeler driving, underscoring the need for datasets and benchmarks tailored to MTW-centric driving in unstructured environments.

Despite growing interest in MTW-centric perception, no benchmark exists for predictive perception on motorized two-wheelers. Existing MTW datasets such as MOTOR~\cite{motor} support rider action and intention recognition but provide no
trajectory annotations for surrounding agents, while trajectory prediction datasets~\cite{chandra2019traphic,meteor} focus on four-wheelers. No dataset jointly supports rider maneuver anticipation and surrounding-agent trajectory prediction from an MTW's ego-centric perspective, or characterizes the riding conditions under which prediction is evaluated.

\noindent To bridge this gap, we introduce \textbf{MotoVerse}, an MTW-centric predictive
benchmark. Our main contributions are as follows:
\begin{itemize}
\item \textbf{MotoVerse Benchmark.} A large-scale ego-centric benchmark for predictive perception on motorized two-wheelers, {curated from over 50 hours of collected riding data}. MotoVerse comprises 949 maneuver clips and dense trajectory annotations for over $31{,}000$ surrounding road-agent instances, including more than $16{,}000$ MTWs, enabling two complementary tasks: \textit{Ego-Rider Maneuver Anticipation (ERMA)} and \textit{Traffic Agent Trajectory Prediction (TATP)}.

\item \textbf{MTW Challenge Taxonomy.} A challenge-aware benchmark annotating three distinctive characteristics of MTW riding: \textit{dynamic ego-motion}, \textit{weak traffic structure} and \textit{contextual occlusion}, enabling fine-grained evaluation of predictive models under realistic riding conditions.

\item \textbf{Comprehensive Benchmarking and Insights.} Strong baselines for both ERMA and TATP across established conventional prediction models, {general-purpose VLMs, and recent driving-specific VLMs and vision-language-action models}, together with a preliminary study of parameter-efficient VLM adaptation, providing the first systematic study of predictive reasoning in MTW-centric unstructured traffic.
\end{itemize}

\section{Related Works}
\label{sec:lit_survey}
\subsection{Datasets}

Advancing predictive perception in autonomous driving requires datasets that jointly capture rider intentions, surrounding agent motion, and their interactions in realistic traffic environments. As summarized in Table~\ref{tab:dataset_comparison}, existing datasets provide only a partial view of this problem and remain inadequate for benchmarking predictive tasks from the perspective of MTWs.

\noindent Early intention prediction datasets such as Brain4Cars~\cite{jain2015car}, HDD~\cite{ramanishka2018toward}, AIDE~\cite{yang2023aide} and DAAD~\cite{wasi2024early} annotate driver maneuvers, gaze and multimodal perception, but are collected exclusively from four-wheelers, and all but DAAD in structured traffic. {More recently, MTW-centric datasets such as MOTOR~\cite{motor} have introduced ego-centric videos of realistic two-wheeler riding in dense, unstructured traffic. These datasets are designed for rider action and intention \emph{recognition}, where the full sequence is observed and classified, whereas ERMA targets \emph{anticipation}, requiring the maneuver to
be inferred from an early portion of the clip. This difference also determines the label space, as anticipation requires maneuver classes with identifiable onsets (Sec.~\ref{sec:task_formulation}). Separately, existing MTW datasets provide no trajectory annotations for surrounding traffic participants and therefore cannot support TATP, nor do they provide labels characterizing the riding conditions under which prediction is evaluated.}

\noindent Consequently, no existing dataset jointly supports ERMA and TATP while accounting for the unique challenges of MTW-centric riding. MotoVerse addresses this gap through dense trajectory annotations, rider maneuver labels and challenge-aware metadata for predictive perception in unstructured two-wheeler
traffic.

\subsection{Predictive Methods}

\noindent \textbf{Intention Prediction.} Early video-based intention prediction methods adapted image recognition architectures to model temporal dynamics. C2D~\cite{C2D} aggregates frame-wise features using 2D convolutions, while
I3D~\cite{I3D} extends convolutional kernels into the temporal dimension for joint spatio-temporal learning. More recent architectures such as
SlowFast~\cite{slowfast} and X3D~\cite{feichtenhofer2020x3dexpandingarchitecturesefficient} improve efficiency through multi-pathway and scalable designs. In the driving domain, maneuver anticipation methods further incorporate temporal models such as LSTMs,
transformers, and multi-view perception to predict future driver actions~\cite{gebert2019end, rong2020driver, ma2023cemformer, wasi2024early,
daad-x}. However, these approaches are primarily developed for four-wheelers equipped with multiple synchronized sensors and cameras. Such hardware
configurations are rarely available on MTWs, where a single forward-facing camera remains the most practical sensing modality, and their effectiveness under MTW-specific challenges has not been systematically evaluated.

\begin{table}[t]
\centering

\caption{\textbf{Comparison of existing intention and trajectory prediction
datasets.} MotoVerse is the only dataset providing both ego-rider intention and
surrounding-agent trajectory annotations alongside MTW-specific challenge labels.
\tg~= available, \tr~= unavailable. \textbf{Ego}: 4W = four-wheeler.
\textbf{Camera}: E = ego, S = static. \textbf{Ego Int.}: intention annotations
for the ego driver or rider; datasets annotating surrounding-agent intentions are
marked \tr. {\textbf{Frame}: coordinate frame of the trajectory
annotations (Img.\ = ego-camera image plane, Metric = world or bird's-eye-view;
-- = none).} \textbf{MTW Lbls.}: MTW-specific challenge labels.}
\label{tab:dataset_comparison}
\scriptsize
\setlength{\tabcolsep}{4pt}
\renewcommand{\arraystretch}{1.15}
\begin{tabular}{|l|c|c|c|c|c|c|}
\hline
\textbf{Dataset} & \textbf{Ego} & \textbf{Camera} & {\textbf{Ego Int.}} & \textbf{Traj. Ann.} & {\textbf{Frame}} & \textbf{MTW Lbls.} \\ \hline
\multicolumn{7}{|l|}{\textit{Structured Traffic}} \\ \hline
JAAD~\cite{rasouli2017they_workshop} & 4W & E & \tr & \tg & {Img.} & \tr \\ \hline
{PIE~\cite{PIETraj}} & {4W} & {E} & \tr & \tg & {Img.} & \tr \\ \hline
{HEV-I~\cite{fvl}} & {4W} & {E} & \tr & \tg & {Img.} & \tr \\ \hline
NGSIM~\cite{NSGIM} & 4W & S & \tr & \tg & {Metric} & \tr \\ \hline
KITTI~\cite{kitti} & 4W & E & \tr & \tg & {Metric} & \tr \\ \hline
Waymo~\cite{waymo_open_dataset} & 4W & E & \tr & \tg & {Metric} & \tr \\ \hline
{Argoverse 2~\cite{argoverse2}} & {4W} & {E} & \tr & \tg & {Metric} & \tr \\ \hline
{nuPlan~\cite{nuplan}} & {4W} & {E} & \tr & \tg & {Metric} & \tr \\ \hline
AIDE~\cite{yang2023aide} & 4W & E & \tg & \tr & {--} & \tr \\ \hline
HDD~\cite{ramanishka2018toward} & 4W & E & \tg & \tr & {--} & \tr \\ \hline
Brain4Cars~\cite{jain2015car} & 4W & E & \tg & \tr & {--} & \tr \\ \hline
\multicolumn{7}{|l|}{\textit{Unstructured Traffic}} \\ \hline
METEOR~\cite{meteor} & 4W & E & \tr & \tg & {Img.} & \tr \\ \hline
TRAF~\cite{chandra2019traphic} & 4W & S & \tr & \tg & {Metric} & \tr \\ \hline
DAAD~\cite{wasi2024early} & 4W & E & \tg & \tr & {--} & \tr \\ \hline
MOTOR~\cite{motor} & MTW & E & \tg & \tr & {--} & \tr \\ \hline
\textbf{MotoVerse} & MTW & E & \tg & \tg & {Img.} & \tg \\ \hline
\end{tabular}
\end{table}

{
\noindent \textbf{Ego-centric Trajectory Prediction.} Beyond world and bird's-eye-view formulations~\cite{argoverse2, nuplan}, a parallel line of work predicts agent trajectories directly in the image plane of an ego-centric camera,
where the output is a future bounding box sequence. Yao \etal~\cite{fvl} introduced this setting for surrounding vehicles; it has since supported accident anticipation~\cite{fol}, uncertainty modeling~\cite{bhattacharyya2018longtermonboardpredictionpeople} and traffic agent forecasting in risky driving scenarios~\cite{fusiongru}. Recent methods such as PIETraj~\cite{PIETraj}, SGNet~\cite{SGNet}, BiTraP~\cite{bitrap}, and
action-aware contrastive learning~\cite{ABC+} capture multimodal futures through intention reasoning, goal conditioning, and stochastic prediction. Existing work in this setting, however, remains confined to four-wheelers and pedestrians, where the camera component of apparent motion is comparatively stable~\cite{fvl}. On MTWs, rider lean, steering corrections and road vibrations continuously alter camera pose (Fig.~\ref{fig:teaser_fig}), so prediction requires separating agent motion from a large, non-stationary ego component.}

\begin{figure*}[h]
    \centering
    \includegraphics[width=\textwidth]{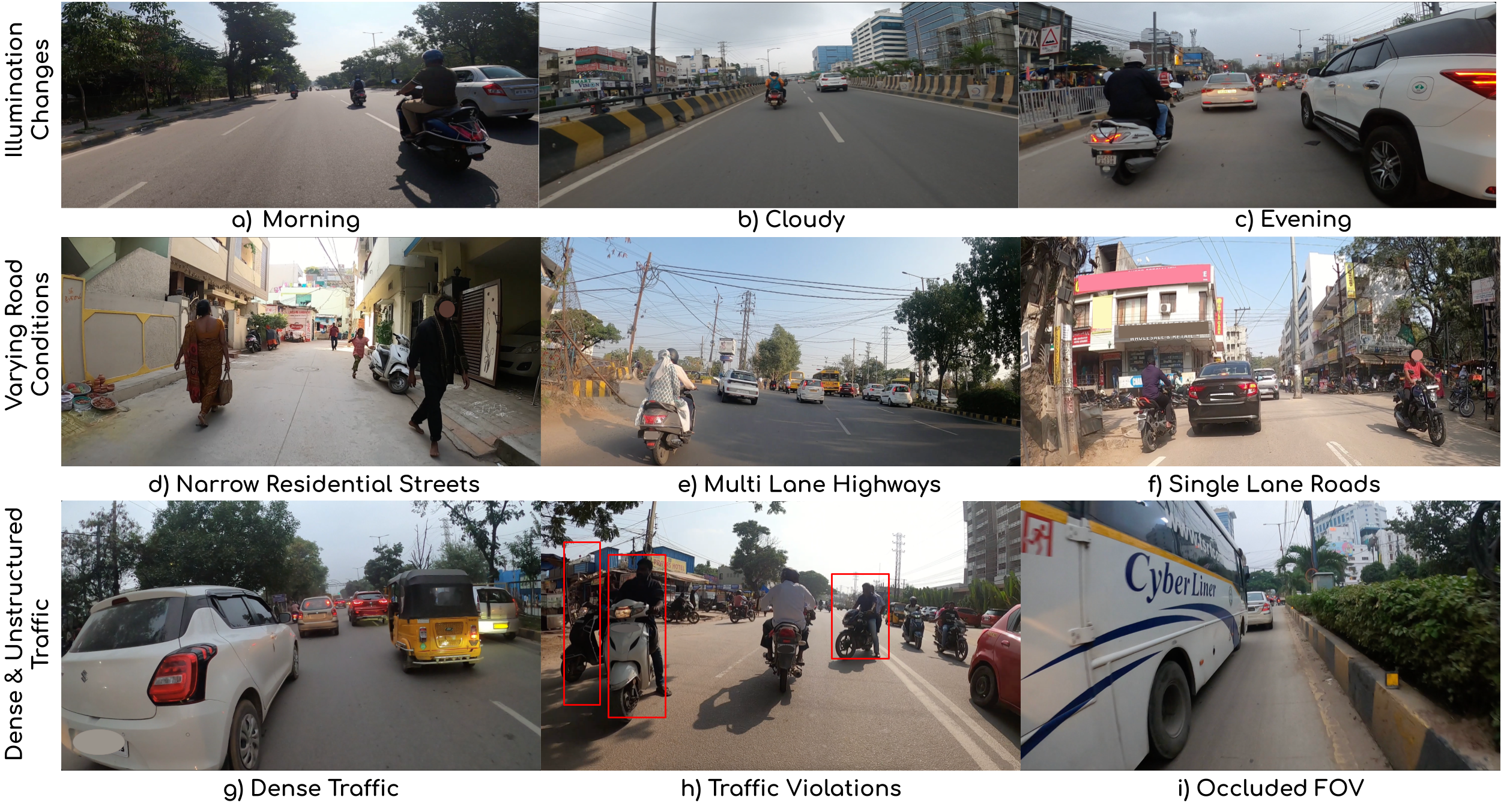}
   \caption{\textbf{Representative samples from the MotoVerse dataset.}
MotoVerse captures the diversity and complexity of MTW-centric driving in dense,
unstructured traffic, including varying illumination conditions (top), diverse
road types from highways to narrow urban streets (middle), and heterogeneous
traffic scenarios characteristic of the Global South (bottom).}
    \label{fig:dataset_samples}
\end{figure*}

\noindent \textbf{VLM-Based Predictions.} Recent VLMs have demonstrated remarkable reasoning capabilities and are increasingly being adopted for autonomous driving tasks including traffic scene understanding, visual question answering, and
decision making, as exemplified by nuScenes-QA~\cite{qian2023nuscenes}, DriveLM~\cite{sima2023drivelm},
CoVLA~\cite{arai2025covlacomprehensivevisionlanguageactiondataset}, and
OpenDriveVLA~\cite{zhou2025opendrivevlaendtoendautonomousdriving}. {Alongside general-purpose models such as InternVL2.5~\cite{InternVL} and Qwen3-VL~\cite{qwen}, driving-specific models
have extended this to explicit reasoning over driving scenes~\cite{cosmos,drivemmo1} and to action prediction~\cite{unidrivevla,alpamayo}.} However, existing evaluations are almost exclusively conducted on four-wheeler datasets collected in structured traffic environments, and their ability to reason about MTW intentions and
surrounding agent behavior in MTW-centric, unstructured traffic remains to be
tested.

\section{The MotoVerse Benchmark}
\label{sec:dataset}

We present MotoVerse, a dataset and benchmark for predictive perception from the ego-centric perspective of MTWs in dense, unstructured traffic. This section describes our data capture setup and collection, the annotation protocol, the quality assurance process, and the resulting dataset statistics and evaluation splits.

\subsection{Data Collection}
MotoVerse was collected using a single forward-facing GoPro Hero~8 camera mounted
on a motorized two-wheeler, recording at 1080p and 30~FPS.
{A single forward camera is the sensing configuration realistically
deployable on production two-wheelers in the Global South
(Sec.~\ref{sec:intro}), and no additional sensors were used. The camera's
internal IMU is used only to illustrate the ego-motion characteristics of MTW
riding in Fig.~\ref{fig:teaser_fig}; both benchmark tasks are defined on the
video stream alone.}
The camera was positioned to capture the rider's frontal field of view,
providing an ego-centric perspective of the surrounding traffic that closely
reflects the visual context available to the rider during navigation.

\noindent The dataset was recorded over more than 50 hours of riding, covering
approximately 700~km across diverse urban and suburban environments
{in India,} in dense, unstructured traffic. The recordings span a
wide range of traffic densities, road types, varied illumination
conditions, and riding scenarios, capturing the heterogeneity of real-world MTW
operation. From this corpus, we curated the clips used to construct the
MotoVerse benchmark for predictive perception.

\noindent\textbf{Ethical Statement.} Prior to data collection, all participants provided informed consent under a protocol reviewed and approved by our Institutional Review Board (IRB). All riders were above 18 years of age and held
valid driving licenses. Recordings were captured under naturalistic riding conditions, with participants briefed on potential risks and supervised throughout each session. The released dataset is fully anonymized and intended strictly for research purposes.

\begin{figure*}[t]
    \centering
    \begin{subfigure}[b]{0.24\linewidth}
        \centering
        \includegraphics[width=\linewidth]{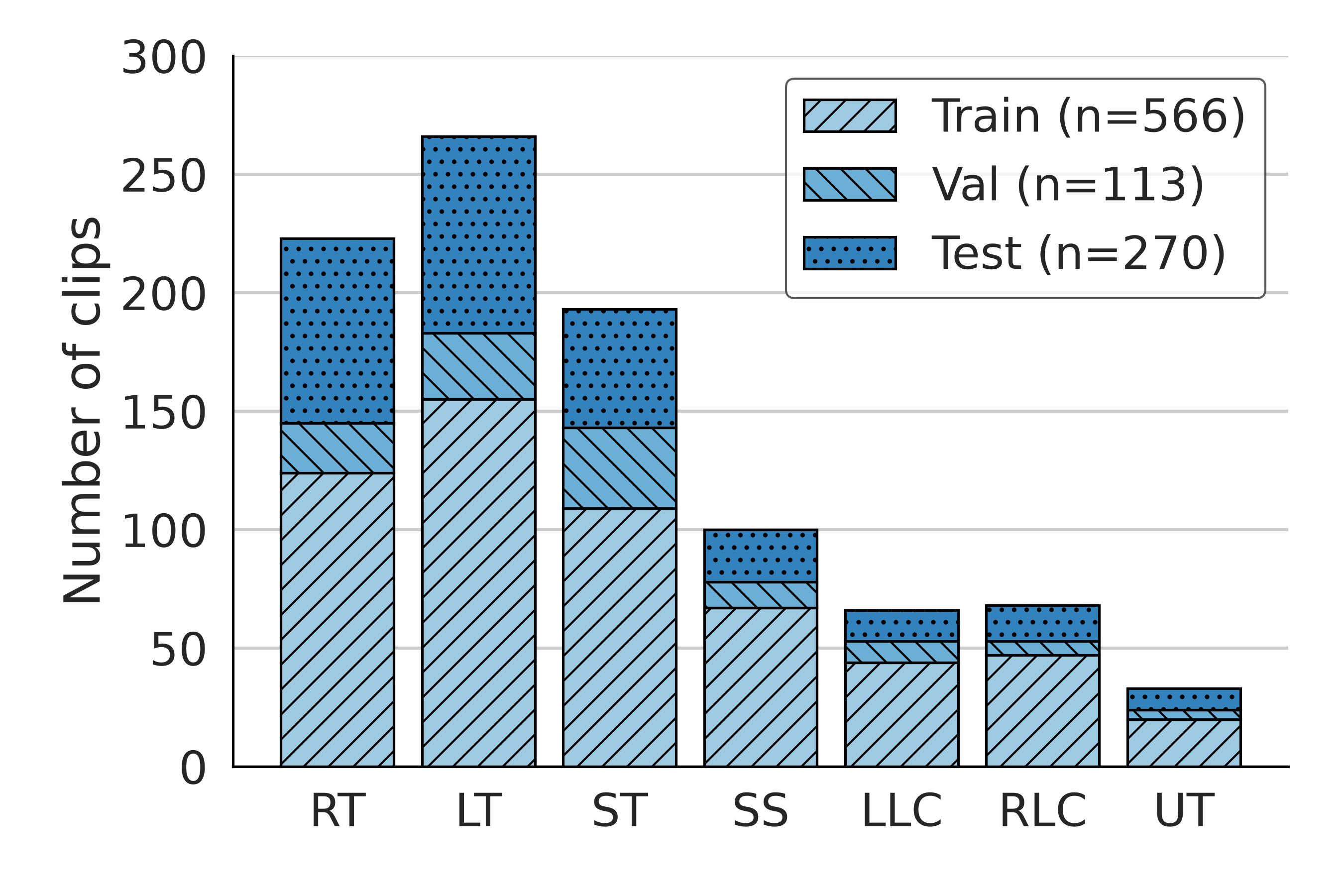}
        \caption{Maneuver Distribution}
        \label{fig:stats_maneuver_dist}
    \end{subfigure}
    \hfill
    \begin{subfigure}[b]{0.24\linewidth}
        \centering
        \includegraphics[width=\linewidth]{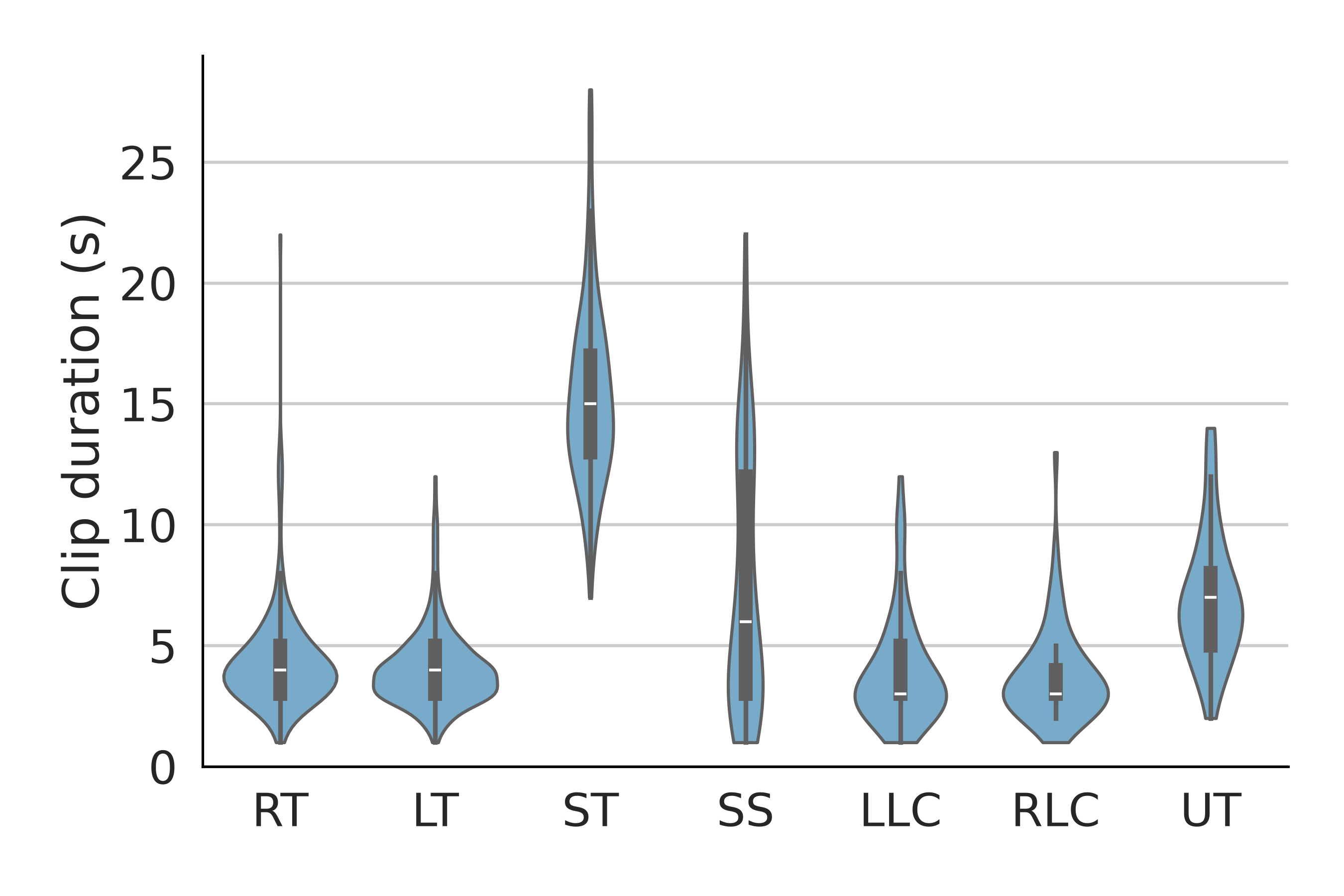}
        \caption{Maneuver Duration}
        \label{fig:stats_maneuver_dur}
    \end{subfigure}
    \hfill
    \begin{subfigure}[b]{0.24\linewidth}
        \centering
        \includegraphics[width=\linewidth]{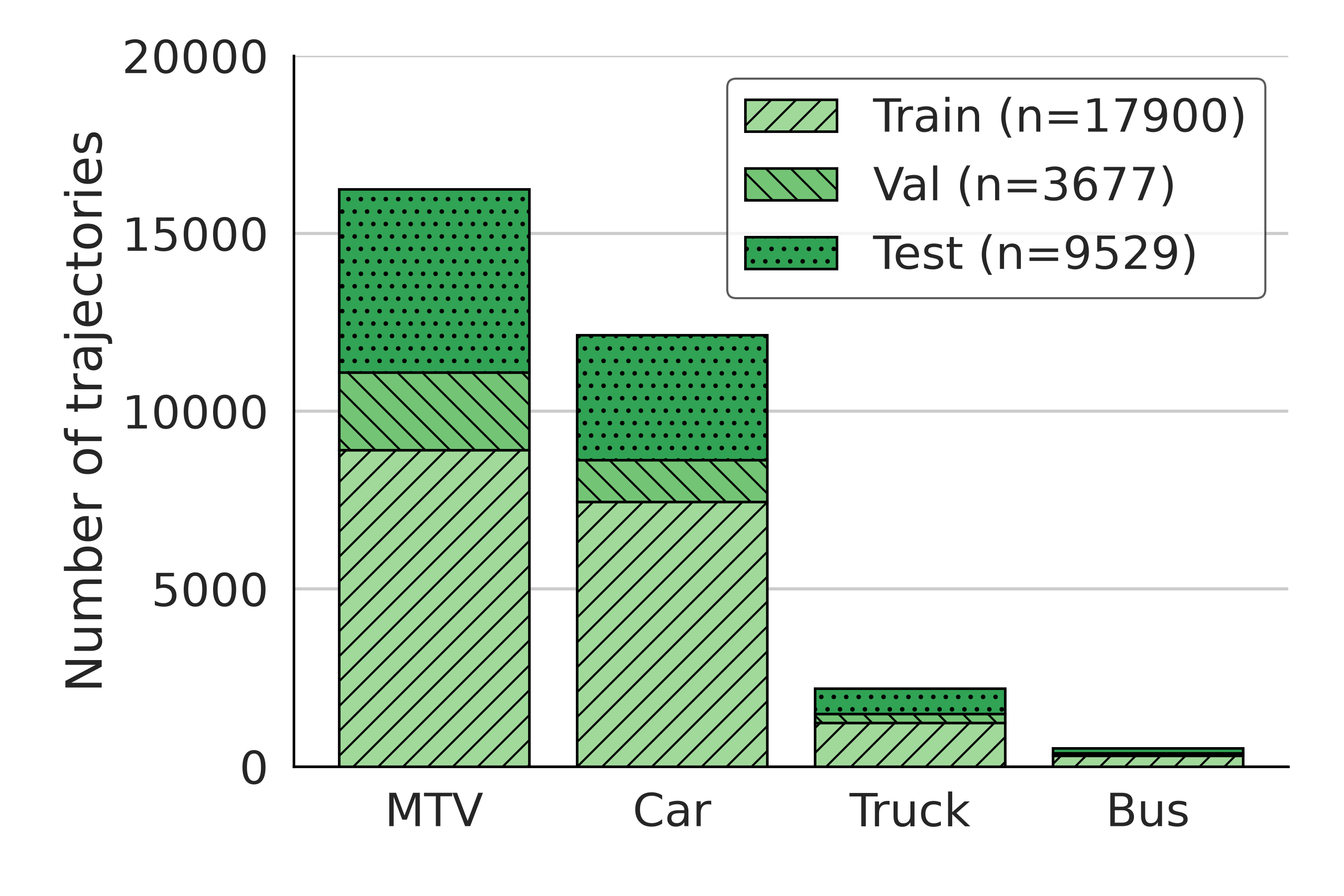}
        \caption{Agent Distribution}
        \label{fig:stats_agent_dist}
    \end{subfigure}
    \hfill
    \begin{subfigure}[b]{0.24\linewidth}
        \centering
        \includegraphics[width=\linewidth]{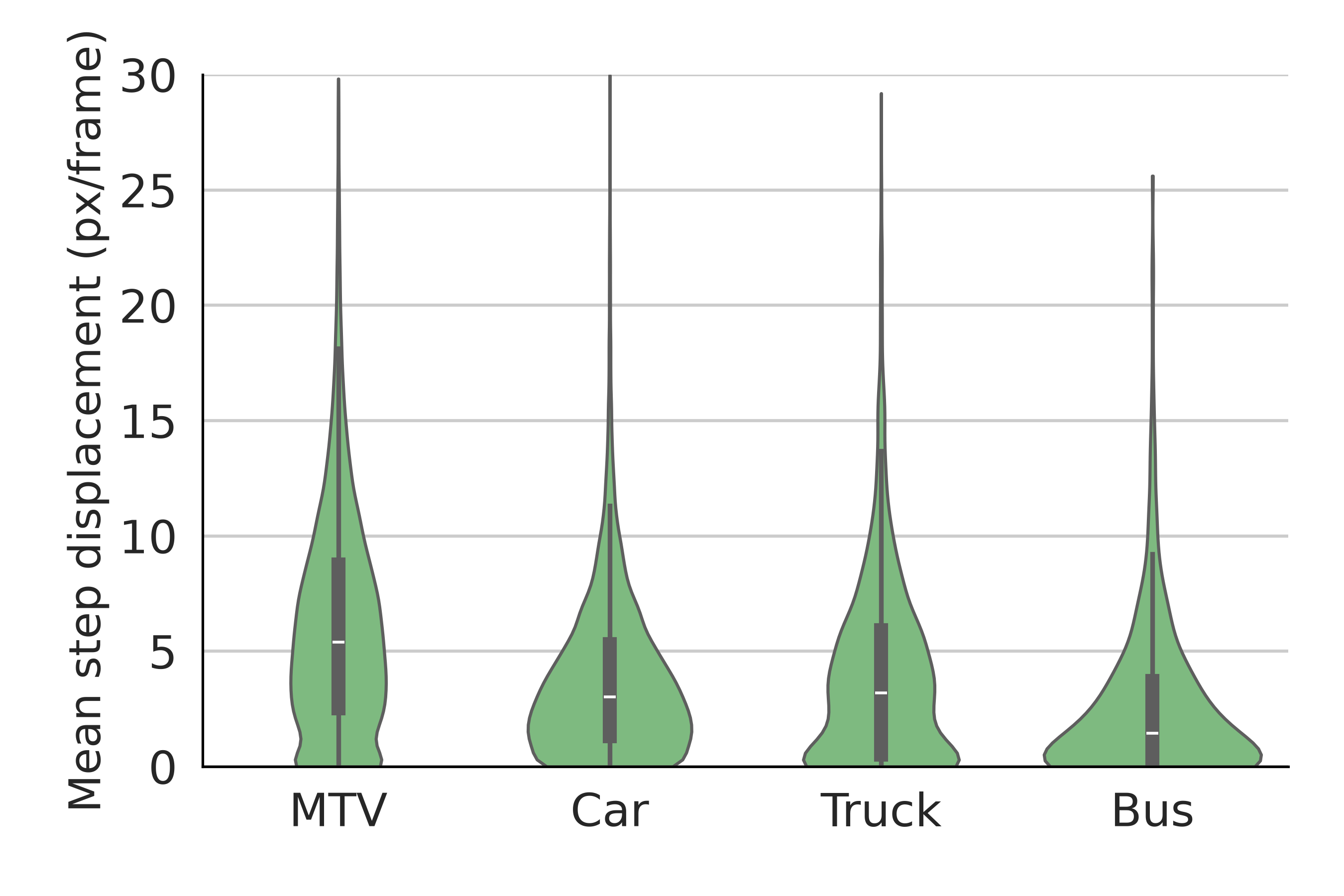}
        \caption{Motion Variance}
        \label{fig:stats_motion_var}
    \end{subfigure}
    
    \caption{\textbf{MotoVerse dataset statistics.} (a) Distribution of ego-rider
    maneuver classes. (b) Distribution of maneuver durations. (c) Distribution of
    surrounding traffic agents. (d) Average frame-to-frame displacement across
    agent categories, highlighting their distinct motion characteristics.
    Abbreviations: \textbf{ST}—Straight, \textbf{LT}—Left Turn, \textbf{RT}—Right
    Turn, \textbf{LLC}—Left Lane Change, \textbf{RLC}—Right Lane Change,
    \textbf{UT}—U-Turn, and \textbf{SS}—Slow/Stop.}
    \label{fig:dataset_stats}
\end{figure*}

\subsection{MotoVerse Dataset}

MotoVerse comprises 949 video clips curated from over 50 hours of riding, each capturing a single ego-rider maneuver, totaling approximately 4.5 hours with an average clip duration of 17 seconds. Every clip carries an ego-rider maneuver label from seven categories, dense frame-level trajectories for the surrounding road agents, and {three binary MTW-specific challenge labels describing dynamic ego-motion, weak traffic structure and contextual occlusion}.
In total, the benchmark provides trajectories for over 31,000 road-agent instances, including more than 16,000 MTWs, alongside cars, buses and trucks
(Fig.~\ref{fig:dataset_samples}).

\subsection{Annotation Pipeline}
\label{annotations}
We first annotated each clip with its ego-rider maneuver, and then annotated the surrounding-agent trajectories on the labeled clips, along with the MTW challenge labels. Maneuvers were grouped into seven standard categories: \textit{Straight (ST), Left Turn (LT), Right Turn (RT), Left Lane Change (LLC), Right Lane Change (RLC), Slow/Stop (SS), and U-Turn (UT)}. To efficiently annotate the surrounding-agent trajectories, we adopted a semi-automated pipeline. Vehicle detections for MTWs, cars, buses, and trucks were first generated using YOLOv11~\cite{yolov11}, followed by trajectory initialization using the ByteTrack multi-object tracker~\cite{zhang2022bytetrack}. Owing to dense interactions and frequent occlusions in unstructured traffic, automatically generated tracks occasionally suffer from fragmented trajectories and identity switches.

\noindent To ensure annotation quality, a team of five annotators, supervised by a domain expert, manually verified and corrected all trajectories before converting them into the JAAD annotation format~\cite{kotseruba2020jointattentionautonomousdriving},
to enable compatibility with existing frameworks. The same team annotated the ego-rider maneuver labels and the video-level MTW challenge labels using predefined annotation guidelines. Specifically, \textbf{Dynamic ego-motion} denotes clips exhibiting pronounced camera motion caused by rider lean, steering, or road vibrations; \textbf{Weak traffic structure} identifies scenes where road agents including the ego vehicle frequently deviate from lane structure through irregular lateral maneuvers; and \textbf{Contextual occlusion} indicates that key traffic participants are substantially obscured by surrounding vehicles. Weak traffic structure was labeled manually. Dynamic ego-motion and contextual occlusion were first scored automatically from camera-motion and occlusion estimates, and every resulting label was then verified by the annotators.
{Each clip receives an independent binary label for each
of the three challenges, so a clip may carry none, one, two, or all three.}

\noindent\textbf{Annotation Quality.} To assess the reliability of the annotations, we measured inter-annotator agreement among the five annotators
using vote variance. The initial annotation process yielded an ambiguity rate of less than 6\%, with disagreements primarily arising in highly occluded or visually challenging scenes. Starting with an initial set of 1,068 annotated
clips, we removed ambiguous and corrupted samples through quality control, yielding a final benchmark of 949 clips. Additional details on the annotation protocol and agreement analysis are provided in the \textit{Supplementary
Material}.

\subsection{Dataset Statistics and Benchmark Splits} 
MotoVerse captures the diversity and complexity of MTW-centric driving in dense and unstructured traffic. As illustrated in Fig.~\ref{fig:dataset_stats}, MTWs and cars constitute the majority of surrounding traffic participants, with MTWs showing the largest average frame-to-frame displacement owing to their agile maneuvers in congested traffic. The maneuver annotations span a wide range of categories and durations, requiring models to anticipate intentions from variable-length observation windows. 

\noindent {Under this taxonomy, 693 clips exhibit dynamic ego-motion, 457
weak traffic structure, and 210 contextual occlusion; as the labels are independent, these counts overlap rather than partition the dataset.} These annotations enable fine-grained robustness analysis under challenging riding conditions. To ensure a rigorous evaluation and eliminate session-level data leakage, we partitioned the benchmark according to rider sessions. Specifically, the dataset comprises 16 independent riding sessions from 16 distinct riders, of which 9 are used for training, 3 for validation and 4 for testing (566, 113 and 270 clips, respectively), ensuring that geographic context and ride-specific characteristics remain disjoint across the three splits.

\subsection{Task Formulation}
\label{sec:task_formulation}
 
{We define both tasks under a monocular vision-only sensing
assumption: models receive a single forward-facing camera stream and no other
input, reflecting the sensing configuration realistically deployable on
production two-wheelers (Sec.~\ref{sec:intro}). Formal input-output specifications, evaluation
protocols and metric definitions are provided in the \textit{Supplementary
Material}.}
 
{\noindent\textbf{ERMA.} Given the first $\tau$ fraction of a clip
containing a single ego-rider maneuver, the model predicts the maneuver class from the seven categories defined in Sec.~\ref{annotations}, following the maneuver taxonomies used in prior driver anticipation work~\cite{jain2015car,ramanishka2018toward,wasi2024early} and adapted to two-wheeler riding.}
 
\noindent {The label space is restricted to discrete actions rather than riding
behaviors, since $\tau$ is defined as the fraction of a maneuver observed and every label therefore requires an identifiable onset. Behaviors defined for MTWs~\cite{motor} such as weaving through traffic do not satisfy this requirement. The same requirement governs segmentation: clips are cut so that each contains exactly one maneuver, and segments spanning more than one maneuver are not retained.}

\noindent\textbf{TATP.} Given the observed bounding-box trajectory of a
surrounding traffic agent, the model predicts its future bounding-box trajectory
in the ego-camera image plane. { The formulation follows the future object
localization setting established in prior ego-centric work~\cite{fvl,fol,fusiongru}, keeping TATP comparable with existing image-plane
forecasting benchmarks, and we adopt the displacement and overlap metrics
standard in that literature~\cite{citywalks,PIETraj}}.

\section{Experiments and Analysis}

We evaluate MotoVerse on ERMA and TATP, benchmarking conventional prediction
models trained on MotoVerse against zero-shot general-purpose and driving-specific VLMs and VLAs, and analyze performance under varying observation horizons and across the proposed MTW challenge taxonomy.

\subsection{Experimental Settings}
\noindent\textbf{Evaluated Methods.} For ERMA, we evaluated four widely used video backbones: C2D~\cite{C2D}, I3D~\cite{I3D}, X3D~\cite{feichtenhofer2020x3dexpandingarchitecturesefficient} and SlowFast~\cite{slowfast}. For TATP, we evaluated PIETraj~\cite{PIETraj}, BiTraP~\cite{bitrap}, ABC+~\cite{ABC+} and SGNet~\cite{SGNet}, along with a static baseline that holds the last observed box fixed. We also evaluated four
general-purpose VLMs, InternVL2.5 (1B, 4B)~\cite{InternVL} and Qwen3-VL (2B, 4B)~\cite{qwen}, and four driving-pretrained models: Cosmos-Reason2-8B~\cite{cosmos}, DriveLMM-o1~\cite{drivemmo1}, UniDriveVLA (Base)~\cite{unidrivevla} and Alpamayo-1.5-10B~\cite{alpamayo}. Unless stated otherwise, all VLMs and VLAs were evaluated zero-shot on both tasks.

\noindent\textbf{Implementation Details.} Each experiment ran on a single
NVIDIA L40S GPU. For a given task, all models received identical visual inputs, and all VLMs identical prompts. Conventional models were trained on the MotoVerse training split with the authors' default hyperparameters; ERMA backbones were trained per $\tau$ for 100 epochs, keeping the checkpoint with the best validation accuracy; TATP models kept the checkpoint with the lowest validation loss. VLMs were run in bf16 with greedy decoding. Further details are in the \textit{Supplementary Material}.

\noindent\textbf{Evaluation Protocol.} ERMA is evaluated at four observation
ratios, $\tau \in \{0.2, 0.3, 0.5, 1.0\}$, using Accuracy (Acc), weighted-F1 and
macro-F1. For TATP, models observe 15 frames (0.5\,s) and predict the next 45
(1.5\,s)~\cite{bitrap}. We report Average and Final Displacement Error
(ADE/FDE), their scale-normalized counterparts (sADE/sFDE), and Average and
Final Bounding-box IoU (AIoU/FIoU). As consecutive windows of a track overlap
heavily, our primary TATP set takes the first window of each continuous test
track (1,246 windows), giving one sample per agent; results on all 10,133
windows, with largely unchanged rankings, are in the \textit{Supplementary
Material}. Unparsed VLM answers are excluded on ERMA, with the parse rate
reported, and scored as the last observed box on TATP.

\subsection{ERMA Results and Analysis}
\label{sec:erma_results}

\begin{table}[t]
\centering
\caption{\textbf{ERMA performance across different observation ratios ($\tau$).}
Each entry reports \emph{$\mathrm{Acc}$ / Weighted-F1 / Macro-F1}.
The \textbf{best} and \underline{second-best} results in each column are
highlighted in bold and underline, respectively{, ranked by
Acc}.}
\label{tab:erma_main}
\scriptsize
\setlength{\tabcolsep}{3pt}
\renewcommand{\arraystretch}{1.1}
\begin{tabular}{l cccc}
\toprule
\textbf{Model} & $\tau{=}0.2$ & $\tau{=}0.3$ & $\tau{=}0.5$ & $\tau{=}1.0$ \\
\midrule
\multicolumn{5}{l}{\textit{Video backbones (fine-tuned)}} \\
\midrule
C2D~\cite{C2D} & 48.1/43.2/38.7 & \underline{48.1/47.1/36.7} & 45.2/44.2/36.6 & 54.1/52.8/49.6 \\
I3D~\cite{I3D} & 48.5/48.7/43.3 & 41.9/39.1/31.1 & \textbf{53.3/50.5/38.7} & 55.9/53.9/47.0 \\
SlowFast~\cite{slowfast} & \textbf{51.1/49.7/43.4} & \textbf{50.0/46.9/38.1} & 46.7/45.0/41.5 & 54.8/52.9/44.5 \\
X3D~\cite{feichtenhofer2020x3dexpandingarchitecturesefficient} & \underline{49.3/46.7/40.7} & 46.3/44.1/42.9 & \underline{48.9/44.7/37.2} & \underline{58.5/53.8/43.2} \\
\midrule
\multicolumn{5}{l}{\textit{Vision-Language Models (zero-shot)}} \\
\midrule
InternVL2.5-1B~\cite{InternVL} & 31.5/17.9/9.2 & 32.2/18.5/9.7 & 32.2/18.5/9.7 & 30.0/15.6/7.6 \\
InternVL2.5-4B~\cite{InternVL} & 30.0/18.7/12.8 & 31.5/17.8/12.1 & 32.2/17.2/11.8 & 31.5/17.7/14.1 \\
Qwen3-VL-2B~\cite{qwen} & 28.5/15.7/10.6 & 28.9/16.1/10.5 & 30.0/16.6/10.8 & 29.3/17.1/11.6 \\
Qwen3-VL-4B~\cite{qwen} & 17.0/13.2/10.1 & 22.6/20.3/15.7 & 34.1/34.1/23.2 & 39.3/39.9/26.0 \\
\midrule
\multicolumn{5}{l}{\textit{Driving-specific VLMs / VLAs (zero-shot)}} \\
\midrule
{Cosmos-Reason2-8B~\cite{cosmos}} & {21.9/14.8/12.4} & {26.7/22.9/18.1} & {40.7/38.6/29.1} & {\textbf{60.0/57.0/35.4}} \\
{UniDriveVLA~\cite{unidrivevla}} & {22.2/16.0/12.8} & {24.8/19.0/16.1} & {33.7/30.1/24.6} & {39.6/36.6/23.0} \\
{DriveLMM-o1~\cite{drivemmo1}} & {21.5/12.3/13.6} & {21.1/11.1/13.5} & {22.6/13.1/14.7} & {23.7/14.4/15.8} \\
{Alpamayo-1.5-10B$^\dagger$~\cite{alpamayo}} & {9.0/10.4/5.8} & {7.5/8.6/5.1} & {9.9/11.9/8.2} & {11.4/13.7/10.7} \\
\bottomrule
\end{tabular}
\par\vspace{2pt}
\parbox{\linewidth}{\scriptsize $^\dagger$Scores over parsed answers; Alpamayo-1.5-10B
returned a parseable answer for 82.2/84.4/86.3/90.7\% of test clips at
$\tau{=}0.2/0.3/0.5/1.0$. All other models parse 100\%.}
\end{table}

\begin{figure*}[t]
    \centering
    \includegraphics[width=0.8\textwidth]{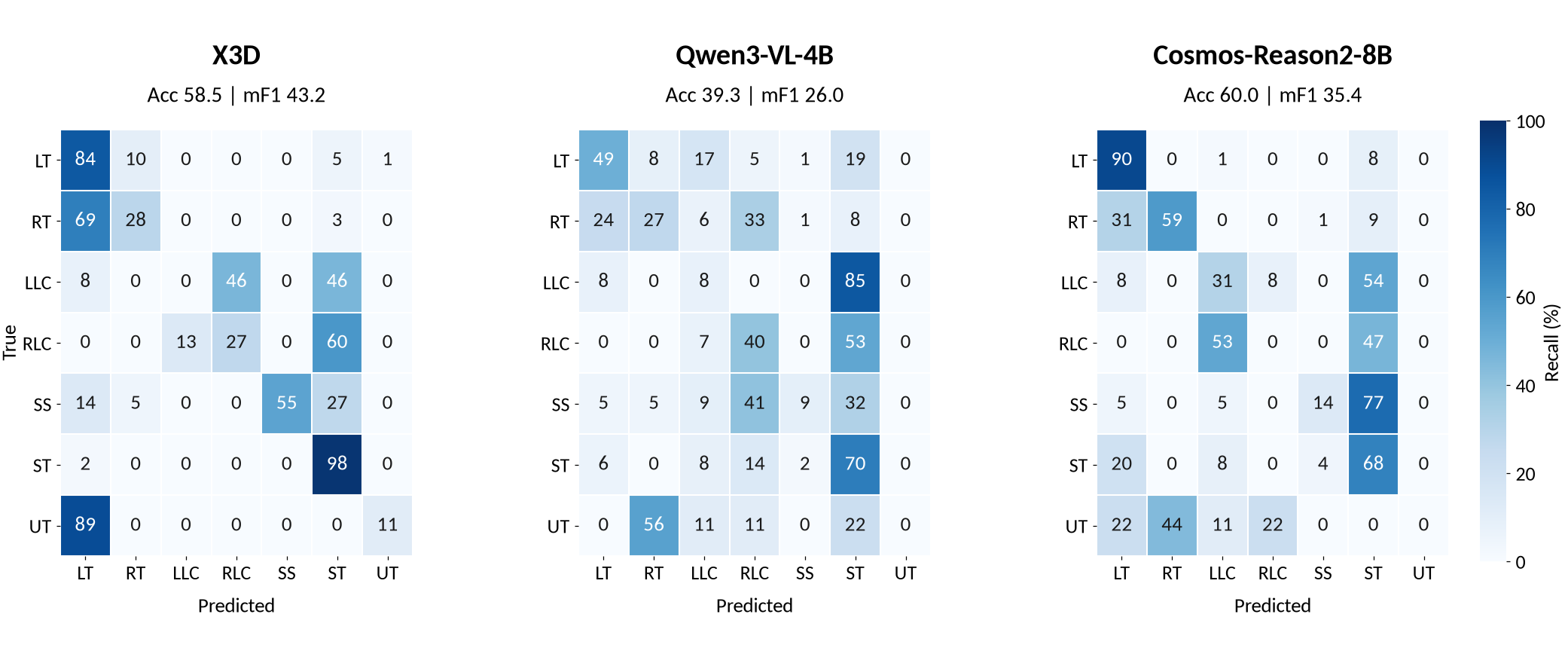}
\caption{\textbf{Row-normalized confusion matrices (\%) at $\tau{=}1.0$} for the
strongest model from each family: a fine-tuned backbone (X3D), a general-purpose VLM (Qwen3-VL-4B), and a driving-specific VLM (Cosmos-Reason2-8B).}
    \label{fig:erma_confusion}
\end{figure*}

\noindent\textbf{Main Results.} Table~\ref{tab:erma_main} reports ERMA performance. The four video backbones perform comparably, and their ranking varies with $\tau$. All clearly exceed the majority-class baseline of 30.7\% (always predicting Left Turn). Their macro-F1 is up to 15 points lower than accuracy, as Right Lane Change and U-Turn are rarely recognized. Even at full observation, the best backbone reaches only 58.5\%. Most zero-shot VLMs perform near or below the majority-class baseline. InternVL2.5 (1B, 4B) and
Qwen3-VL-2B stay near 30\% at every $\tau$, predicting only one or two classes. Qwen3-VL-4B, Cosmos-Reason2-8B and UniDriveVLA improve with longer observation. Cosmos-Reason2-8B reaches 60.0\% at $\tau{=}1.0$, exceeding all video backbones in accuracy, but its macro-F1 (35.4) is well below theirs. DriveLMM-o1 and Alpamayo-1.5-10B predict mostly Straight and U-Turn, respectively.

\noindent\textbf{Effect of $\tau$.} The video backbones change little between $\tau{=}0.2$ and $\tau{=}0.5$ and gain mainly at full observation. At $\tau{=}0.2$, each already reaches 84--93\% of its full-observation accuracy. Predicting from the first fifth of a maneuver thus costs only 4--9 points, while leaving most of the maneuver for a downstream ADAS response. Early accuracy itself, however, remains near 50\%. The zero-shot VLMs behave differently. Qwen3-VL-4B, Cosmos-Reason2-8B and UniDriveVLA gain most after
$\tau{=}0.3$, once the maneuver is under way. They recognize maneuvers rather than anticipate them.

\noindent\textbf{Confusion Matrix Analysis.} {Fig.~\ref{fig:erma_confusion}
compares the error structure of the strongest model from each family at full observation ($\tau{=}1.0$). All three struggle most with the lane-change classes and U-Turns. Lane changes are mostly predicted as Straight (46--85\%) or assigned the wrong direction. A likely cause is that, from an MTW-mounted camera, a lane change is a small lateral shift that resembles riding straight. U-Turn is recognized in at most 11\% of cases. X3D is biased towards Left Turn, which absorbs 69\% of Right Turn and 89\% of U-Turn clips. Cosmos-Reason2-8B separates the two turns better (59\% vs.\ 28\% Right Turn recall). However, it
assigns 77\% of Slow/Stop clips to Straight and never recognizes Right Lane Change or U-Turn. Qwen3-VL-4B often predicts the right direction but the wrong maneuver, labeling 33\% of Right Turn clips as Right Lane Change. Matrices for all models across observation ratios are in the \textit{Supplementary Material}.}

\begin{table}[!ht]
\centering
\caption{\textbf{TATP performance comparison.} Stochastic methods report the
mean over $K$ sampled trajectories; deterministic models and VLMs report a single trajectory. VLMs are evaluated zero-shot. The \textbf{best} and
\underline{second-best} results per metric are shown in bold and underline.
{A dash (--) denotes that no parseable trajectory was produced.}}
\label{tab:tatp_main}
\scriptsize
\setlength{\tabcolsep}{3pt}
\begin{tabular}{l cccc cc}
\toprule
\textbf{Model} & ADE\,$\downarrow$ & FDE\,$\downarrow$ & sADE\,$\downarrow$
& sFDE\,$\downarrow$ & AIoU\,$\uparrow$ & FIoU\,$\uparrow$ \\
\midrule
Static (last box held) & 94.3 & 193.4 & 0.705 & 1.259 & 0.363 & 0.151 \\
\midrule
\multicolumn{7}{l}{\textit{Conventional methods (fine-tuned)}} \\
\midrule
PIETraj~\cite{PIETraj} & 73.3 & 145.4 & 0.566 & 0.971 & 0.405 & 0.236 \\
BiTraP~\cite{bitrap}  & \underline{55.5} & \underline{132.7} & \underline{0.423} & \underline{0.963} & \underline{0.528} & 0.249 \\
ABC+~\cite{ABC+}    & 58.5 & 138.3 & 0.474 & 1.046 & 0.520 & \underline{0.252} \\
SGNet~\cite{SGNet}   & \textbf{51.8} & \textbf{115.0} & \textbf{0.389} & \textbf{0.785} & \textbf{0.549} & \textbf{0.312} \\
\midrule
\multicolumn{7}{l}{\textit{Vision-Language Models (zero-shot)}} \\
\midrule
InternVL2.5-1B~\cite{InternVL} & 103.4 & 194.2 & 0.784 & 1.274 & 0.287 & 0.139 \\
InternVL2.5-4B~\cite{InternVL} & 85.6 & 179.1 & 0.623 & 1.167 & 0.371 & 0.150 \\
Qwen3-VL-2B~\cite{qwen} & 91.7 & 187.2 & 0.690 & 1.247 & 0.370 & 0.156 \\
Qwen3-VL-4B~\cite{qwen} & 70.0 & 159.1 & 0.523 & 1.077 & 0.429 & 0.168 \\
\midrule
\multicolumn{7}{l}{\textit{Driving-specific VLMs / VLAs (zero-shot)}} \\
\midrule
{Cosmos-Reason2-8B~\cite{cosmos}} & {85.8} & {177.1} & {0.703} & {1.255} & {0.401} & {0.169} \\
{UniDriveVLA~\cite{unidrivevla}} & {95.1} & {192.3} & {0.693} & {1.229} & {0.354} & {0.147} \\
{DriveLMM-o1~\cite{drivemmo1}} & {89.1} & {184.7} & {0.656} & {1.190} & {0.371} & {0.152} \\
{Alpamayo-1.5-10B~\cite{alpamayo}} & {--} & {--} & {--} & {--} & {--} & {--} \\
\bottomrule
\end{tabular}
\end{table}

\subsection{TATP Results and Analysis}
\label{tatp}

\noindent\textbf{{Main Results.}} Table~\ref{tab:tatp_main} reports TATP
performance. Among the conventional methods, SGNet performs best on every
metric, followed by BiTraP and ABC+. This ranking holds for both ADE and sADE.
All four methods clearly outperform the static baseline, with SGNet lowering
ADE by 45\%. Most zero-shot VLMs improve little over the static baseline, and
InternVL2.5-1B is worse on every metric. Qwen3-VL-4B is the strongest VLM. It
outperforms PIETraj on ADE, sADE and AIoU, but not on the final-step metrics.
All VLMs remain well behind SGNet and BiTraP. Alpamayo-1.5-10B answers almost
entirely in prose and produces no parseable trajectories (examples in the
\textit{Supplementary Material}).

\begin{figure}[!ht]
    \centering
    \includegraphics[width=\columnwidth]{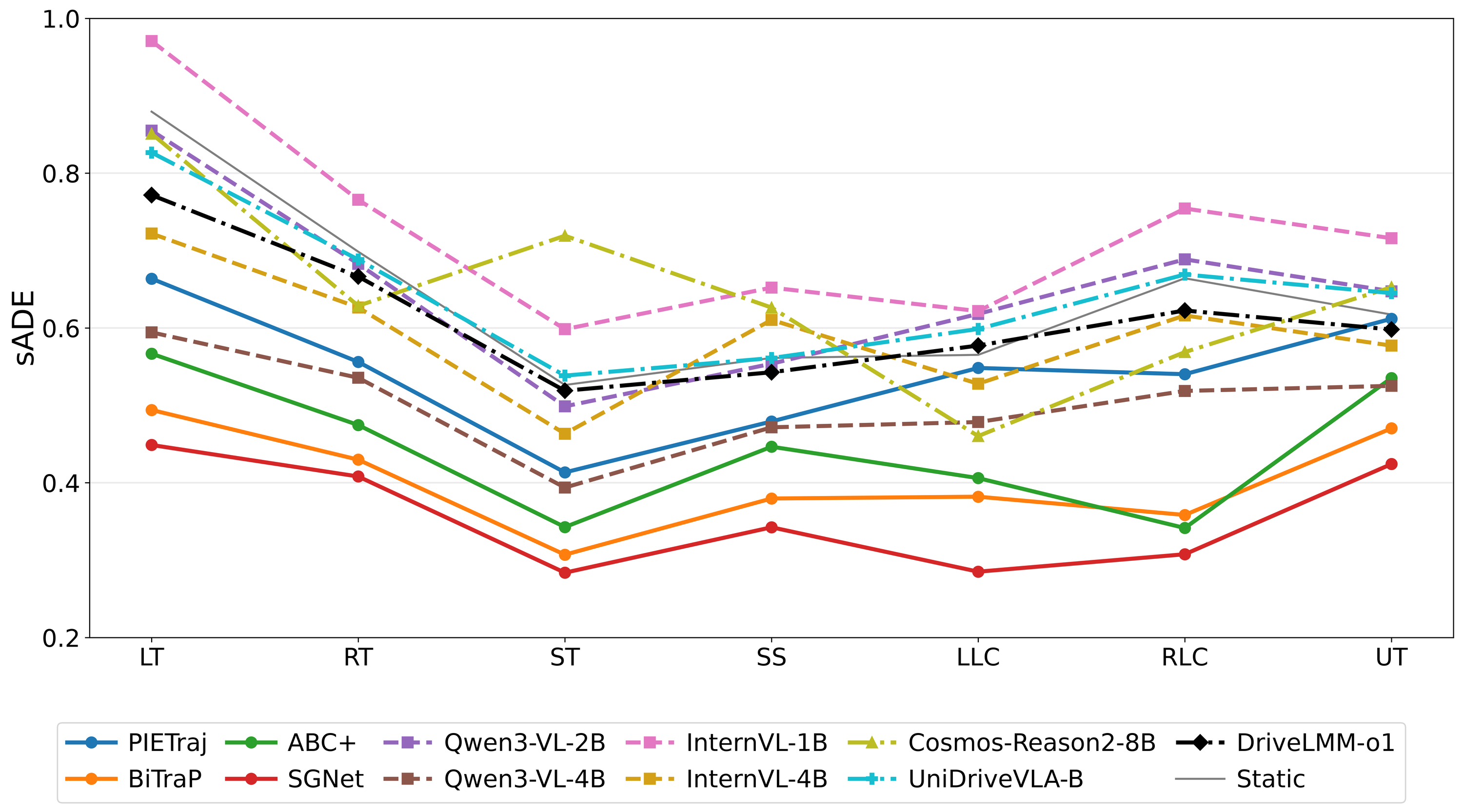}
    \caption{\textbf{Per-maneuver TATP performance (sADE, $\downarrow$).} Results are grouped by the ego-rider maneuver associated with each source clip. The
    grey line shows the static baseline.}
    \label{fig:tatp_maneuver}
\end{figure}

\noindent\textbf{Per-Maneuver Analysis.} Fig.~\ref{fig:tatp_maneuver} shows
trajectory prediction performance across ego-rider maneuvers. {All models
share a similar difficulty profile. Left Turn is the hardest maneuver for
every model, and Straight is the easiest for nearly all. The static baseline
follows the same profile, so difficulty is driven by the ego-rider maneuver
rather than by the model. Cosmos-Reason2-8B is the main exception, with its
second-highest error on Straight. SGNet achieves the lowest error on all seven
maneuvers.}

\subsection{Challenge-wise Robustness Analysis}
\label{sec:challenges}

\noindent\textbf{ERMA.} Table~\ref{tab:erma_challenge} evaluates ERMA under the proposed challenge taxonomy at $\tau{=}0.5$. {The challenge effects are
modest. The fine-tuned backbones lose up to 5 points under weak traffic structure and dynamic ego-motion. Under contextual occlusion, three of the four even
score above their full-set accuracy. I3D performs best on every subset. Contextual occlusion lowers the accuracy of every general-purpose VLM, most
sharply for Qwen3-VL-4B (34.1\% to 19.7\%). Most other VLM scores change by only a few points across subsets. Since many of these models default to one or
two classes, this stability does not reflect robustness. Alpamayo-1.5-10B stays near 10\% throughout.}

\begin{table}[t]
\centering
\caption{\textbf{ERMA $\mathrm{Acc}$ (\%) across MTW-specific challenge subsets}
at $\tau=0.5$. {Full = all test clips, WTS = weak traffic structure, CO = contextual occlusion,
DEM = dynamic ego-motion.} The \textbf{best} and \underline{second-best} results
in each column are highlighted in bold and underline, respectively.}
\label{tab:erma_challenge}
\footnotesize
\setlength{\tabcolsep}{5pt}
\begin{tabular}{l cccc}
\toprule
\textbf{Model} & {Full} & {WTS} & {CO} & {DEM} \\
\midrule
\multicolumn{5}{l}{\textit{Video backbones (fine-tuned)}} \\
\midrule
C2D~\cite{C2D}      & 45.2 & 39.8 & 42.6 & 40.9 \\
I3D~\cite{I3D}      & \textbf{53.3} & \textbf{51.6} & \textbf{57.4} & \textbf{49.2} \\
SlowFast~\cite{slowfast} & 46.7 & 43.0 & \textbf{57.4} & \underline{46.4} \\
X3D~\cite{feichtenhofer2020x3dexpandingarchitecturesefficient}     & \underline{48.9} & \underline{45.3} & \underline{52.5} & 44.2 \\
\midrule
\multicolumn{5}{l}{\textit{Vision-Language Models (zero-shot)}} \\
\midrule
InternVL2.5-1B~\cite{InternVL} & 32.2 & 29.7 & 27.9 & 32.6 \\
InternVL2.5-4B~\cite{InternVL} & 32.2 & 33.6 & 27.9 & 33.1 \\
Qwen3-VL-2B~\cite{qwen} & 30.0 & 33.6 & 24.6 & 32.0 \\
Qwen3-VL-4B~\cite{qwen} & 34.1 & 33.6 & 19.7 & 35.9 \\
\midrule
\multicolumn{5}{l}{\textit{Driving-specific VLMs / VLAs (zero-shot)}} \\
\midrule
{Cosmos-Reason2-8B~\cite{cosmos}} & {40.7} & {35.2} & {39.3} & {42.5} \\
{UniDriveVLA~\cite{unidrivevla}} & {33.7} & {32.8} & {32.8} & {34.3} \\
{DriveLMM-o1~\cite{drivemmo1}} & {22.6} & {25.0} & {23.0} & {23.8} \\
{Alpamayo-1.5-10B$^\dagger$~\cite{alpamayo}} & {9.9} & {9.6} & {13.2} & {7.5} \\
\bottomrule
\end{tabular}
\par\vspace{2pt}
\parbox{\linewidth}{\scriptsize $^\dagger$Over parsed answers only (233 of 270
test clips).}
\end{table}

\noindent\textbf{TATP.} Table~\ref{tab:tatp_challenge} reports trajectory
prediction under the same challenge conditions. Dynamic ego-motion raises sADE
for every model. Weak traffic structure and contextual occlusion lower it for
every model. The static baseline shows the same pattern. Its error measures
how far agents move relative to their size. These differences therefore come
from how much agents move in each subset, not from the models: agents move
more in the image under dynamic ego-motion, and less in the other two subsets.
The ranking is unchanged across subsets, with SGNet best and BiTraP second.
Most zero-shot VLMs stay close to the static baseline on every subset.

\subsection{Exploratory Study: Parameter-Efficient Adaptation of VLMs}
\label{sec:lora}
As a preliminary study, we test whether the zero-shot gap reflects missing adaptation. We fine-tune two VLMs with LoRA~\cite{lora}: Qwen3-VL-2B~\cite{qwen} and UniDriveVLA~\cite{unidrivevla}. UniDriveVLA is
built on Qwen3-VL-2B with the same architecture, so the two models differ mainly in driving pretraining. Training details are in the \textit{Supplementary Material}.

On ERMA, LoRA raises accuracy at $\tau{=}0.5$ from 30.0\% to 67.8\% for Qwen3-VL-2B and from 33.7\% to 76.3\% for UniDriveVLA. Both exceed the best video backbone (53.3\%) and keep improving with longer observation, reaching
80.7\% and 85.9\% at $\tau{=}1.0$. For early anticipation, however, adaptation brings these 2B-parameter VLMs only to the level of the much smaller backbones: at $\tau{=}0.2$, neither exceeds the best backbone (51.1\%). On TATP, LoRA lowers ADE from 91.7 to 58.7 and from 95.1 to 59.6. This only matches ABC+ (58.5) and remains behind BiTraP (55.5) and SGNet (51.8). After adaptation, UniDriveVLA leads Qwen3-VL-2B by 4--9 points on ERMA, while the two are comparable on TATP. Adaptation alone therefore leaves clear headroom on both tasks.

\begin{table}[t]
\centering
\caption{\textbf{TATP sADE, $\downarrow$
across MTW-specific challenge subsets.} {Full = all evaluated windows,
WTS = weak traffic structure, CO = contextual occlusion, DEM = dynamic
ego-motion.} The \textbf{best} and \underline{second-best} results in each
column are highlighted in bold and underline, respectively.}
\label{tab:tatp_challenge}
\footnotesize
\setlength{\tabcolsep}{5pt}
\begin{tabular}{l cccc}
\toprule
\textbf{Model} & {Full} & {WTS} & {CO} & {DEM} \\
\midrule
Static (last box held) & 0.705 & 0.672 & 0.601 & 0.763 \\
\midrule
\multicolumn{5}{l}{\textit{Conventional methods (fine-tuned)}} \\
\midrule
PIETraj~\cite{PIETraj} & 0.566 & 0.540 & 0.492 & 0.598 \\
BiTraP~\cite{bitrap} & \underline{0.423} & \underline{0.401} & \underline{0.371} & \underline{0.439} \\
ABC+~\cite{ABC+} & 0.474 & 0.463 & 0.417 & 0.494 \\
SGNet~\cite{SGNet} & \textbf{0.389} & \textbf{0.372} & \textbf{0.338} & \textbf{0.402} \\
\midrule
\multicolumn{5}{l}{\textit{Vision-Language Models (zero-shot)}} \\
\midrule
InternVL2.5-1B~\cite{InternVL} & 0.784 & 0.726 & 0.671 & 0.829 \\
InternVL2.5-4B~\cite{InternVL} & 0.623 & 0.612 & 0.537 & 0.670 \\
Qwen3-VL-2B~\cite{qwen} & 0.690 & 0.653 & 0.579 & 0.752 \\
Qwen3-VL-4B~\cite{qwen} & 0.523 & 0.496 & 0.436 & 0.547 \\
\midrule
\multicolumn{5}{l}{\textit{Driving-specific VLMs / VLAs (zero-shot)}} \\
\midrule
{Cosmos-Reason2-8B~\cite{cosmos}} & {0.703} & {0.697} & {0.597} & {0.757} \\
{UniDriveVLA~\cite{unidrivevla}} & {0.693} & {0.657} & {0.598} & {0.733} \\
{DriveLMM-o1~\cite{drivemmo1}} & {0.656} & {0.633} & {0.554} & {0.706} \\
\bottomrule
\end{tabular}
\end{table}

\section{Conclusion}
\label{sec:conclusion}

We introduced MotoVerse, a benchmark for predictive perception from the
ego-centric perspective of motorized two-wheelers in dense, unstructured
traffic. MotoVerse jointly supports ERMA and TATP through maneuver, trajectory
and MTW-specific challenge annotations, with strong baselines established using
{conventional prediction models, general-purpose VLMs, and recent
driving-specific VLMs and VLAs}.
Our experiments reveal three key insights. First, {video backbones reach most
of their full-observation accuracy from the first fifth of a maneuver},
providing valuable reaction time for downstream ADAS, whereas zero-shot VLMs
need most of the maneuver to be observed. Second, {zero-shot VLMs, whether
general-purpose or driving-specific, largely remain close to simple baselines
on both tasks, and a preliminary LoRA study shows that adaptation lifts
anticipation sharply but leaves trajectory prediction behind conventional
models}. Third, challenge-aware evaluation indicates that {the rider's own
motion drives forecasting difficulty: for every model, including a static
baseline, errors are highest during left turns and rise under dynamic
ego-motion}. We believe MotoVerse will
accelerate research on safer and more robust predictive AI for intelligent two-wheelers.

\section*{Acknowledgments}
The project was supported by iHub-Data and Mobility at IIIT
Hyderabad. We thank
Prasidh Srikumar and Saumya Mundra for helpful discussions and feedback.

{
    \small
    \bibliographystyle{ieeenat_fullname}
    \bibliography{main}
}

\clearpage
\appendix
\maketitlesupplementary

\section{Dataset Curation and Annotation Quality}

MotoVerse was curated from over 50 hours of ego-centric riding data collected across 16 independent riding sessions. From this corpus, we first segmented the recordings into an initial set of 1,068 clips, each containing a single ego-rider maneuver. {Segments spanning more than one maneuver were not retained, so every clip in the initial set corresponds to exactly one maneuver, as required by the anticipation setting defined in Sec.~3.5 of the main paper.} The temporal boundaries of every maneuver were manually identified by a team of five trained annotators with prior experience in traffic-scene annotation. To ensure annotation consistency, all annotators followed a common
guideline focusing on the maneuver duration, rider behavior, surrounding traffic interactions, and road structure.

We subsequently performed a multi-stage quality control process to construct the final benchmark. First, we removed 42 clips corresponding to Straight and Slow/Stop maneuvers recorded in sparse residential environments containing no
surrounding traffic agents, as these clips provide limited value for either Ego-Rider Maneuver Anticipation (ERMA) or Traffic Agent Trajectory Prediction (TATP). We further discarded 18 clips affected by corrupted frames or severe frame drops that could compromise trajectory annotation quality. Finally, clips with ambiguous maneuver boundaries or labels were independently reviewed by all annotators. Disagreements were resolved through majority voting under the supervision of a domain expert. A total of 59 clips (5.5\%) were identified as ambiguous and removed from the benchmark. {Removal targeted label ambiguity rather than scene difficulty, and the retained benchmark preserves the full range of MTW-specific challenges.} After quality control, the final MotoVerse benchmark comprises 949 clips. The final benchmark contains only consensus annotations, ensuring high-quality labels for both maneuver anticipation and trajectory prediction tasks.

\section{Implementation Details}
\label{sec:supp_impl}

\subsection{Formal Task Definitions}

\subsubsection{Ego-Rider Maneuver Anticipation (ERMA)}

\noindent\textbf{Task Definition.}
Given an ego-centric video clip containing a single rider maneuver, the goal is to predict the maneuver class using only an early portion of the observed video. Formally, let a clip be represented by the frame sequence
$V=\{f_1,\ldots,f_T\}$ with maneuver label
$y\in\{1,\ldots,7\}$. For an observation ratio
$\tau\in(0,1]$, the model observes only the first
$\lceil\tau T\rceil$ frames,
\begin{equation}
V_{\tau}=\{f_1,\ldots,f_{\lceil\tau T\rceil}\},
\end{equation}
and predicts the maneuver as
\begin{equation}
\hat{y}=\arg\max_{c}p(c|V_{\tau}).
\end{equation}

Smaller observation ratios correspond to earlier anticipation with less visual evidence, while $\tau=1.0$ reduces the task to full-video maneuver recognition. MotoVerse defines seven maneuver classes: Left Turn (LT), Right Turn (RT), Left
Lane Change (LLC), Right Lane Change (RLC), Slow/Stop (SS), Straight (ST), and U-Turn (UT){, following the maneuver taxonomies used in prior driver anticipation work~\cite{jain2015car,ramanishka2018toward,wasi2024early} and adapted to two-wheeler riding
as described in Sec.~3.5 of the main paper}.

\noindent\textbf{Evaluation Protocol.}
We evaluate anticipation at four observation ratios,
$\tau\in\{0.2,0.3,0.5,1.0\}$.
Following the rider-session-disjoint protocol described in the main paper, the benchmark contains 566 training, 113 validation and 270 test clips, ensuring that no riding session appears in more than one split. For VLMs, accuracy and F1 scores are computed over parsed answers, and the parse rate is reported.

\begin{table}[h]
\centering
\caption{\textbf{ERMA input/output specification.}}
\label{tab:supp_erma_io}
\footnotesize
\setlength{\tabcolsep}{4pt}
\begin{tabular}{p{1.4cm} p{2.9cm} p{2.9cm}}
\toprule
& \textbf{Video backbones} & \textbf{Vision-Language Models} \\
\midrule
\textbf{Input} & 16 frames sampled within the first-$\tau$ window, RGB & 16 frames sampled within the first-$\tau$ window and a text prompt listing the 7 classes \\
\addlinespace
\textbf{Output} & 7-way softmax over maneuver classes & A single class token, parsed to one of the 7 labels \\
\bottomrule
\end{tabular}
\end{table}

\noindent\textbf{Metrics.} We report Accuracy, weighted-F1, and macro-F1.
Accuracy is the fraction of correctly anticipated maneuvers over the $N$ test
clips,
\begin{equation}
\text{Acc} = \frac{1}{N}\sum_{i=1}^{N} \mathbb{I}(\hat{y}_i = y_i),
\end{equation}
where $y_i$ and $\hat{y}_i$ are the ground-truth and predicted classes for the
$i$-th clip and $\mathbb{I}(\cdot)$ is the indicator function. Given the
per-class precision $\text{P}_c$ and recall $\text{R}_c$ over the $C = 7$
classes, the per-class F1 score is $\text{F1}_c = 2\,\text{P}_c\text{R}_c /
(\text{P}_c + \text{R}_c)$. We aggregate these into weighted-F1 and macro-F1,
\begin{equation}
\text{F1}_w = \sum_{c=1}^{C} w_c\,\text{F1}_c,
\qquad
\text{F1}_m = \frac{1}{C}\sum_{c=1}^{C} \text{F1}_c,
\end{equation}
where $w_c$ is the proportion of true instances in class $c$. Weighted-F1 scales
each class by its frequency, whereas macro-F1 weights every maneuver equally,
providing a balanced assessment under the dataset's class imbalance.

\subsubsection{Traffic Agent Trajectory Prediction (TATP)}

\noindent\textbf{Task Definition.} Given the observed trajectory of a surrounding traffic agent, the goal is to predict its future bounding-box trajectory from the ego-centric perspective of the motorized two-wheeler. For each agent track, the model observes the past trajectory
\begin{equation}
O=\{b_{t-14},...,b_{t}\},
\end{equation}
consisting of the previous 15 bounding boxes, and predicts the future trajectory
\begin{equation}
\hat{P}=\{\hat{b}_{t+1},...,\hat{b}_{t+45}\},
\end{equation}
where each bounding box is represented as $b=[c_x,c_y,w,h]$, corresponding to its center coordinates, width, and height in the image plane{, following the future object localization setting introduced in prior egocentric work~\cite{fvl,fol}}. All coordinates are expressed in pixel space for frames of resolution $960\times540$, {obtained by downsampling the
$1920\times1080$ recordings}. At 30 FPS, the observation horizon corresponds to 0.5 s and the prediction horizon to 1.5 s. The benchmark includes all annotated road-agent categories, namely motorized two-wheelers (MTWs), cars, buses, and trucks.

\noindent\textbf{Evaluation Protocol.} Training and evaluation samples are generated using a sliding-window strategy with a stride of 7 frames over gap-free track segments, with a minimum track length of 61 frames (15 observed frames, 45 prediction frames, and one transition frame){,
following the observation and prediction horizons used in prior
work~\cite{bitrap}}. Under the rider-session-disjoint split described in the main paper, this results in 21,805 training, 3,827 validation and 10,133 test windows. The primary evaluation set used in Tables 3 and 5 and Fig.~5 of the main paper takes the first window of each of the 1,246 continuous agent tracks in the test set, giving one sample per agent. On this subset, the conventional methods keep their full-test-set ranking on ADE, FDE, sADE and AIoU, and the ADE of every method is about 10\% higher (Table~\ref{tab:supp_tatp_full}). Unparsed VLM answers are scored as the last observed box.

\begin{table}[h]
\centering
\caption{\textbf{TATP input/output specification.}}
\label{tab:supp_tatp_io}
\footnotesize
\setlength{\tabcolsep}{4pt}
\begin{tabular}{p{1.4cm} p{2.9cm} p{2.9cm}}
\toprule
& \textbf{Conventional models} & \textbf{Vision-Language Models} \\
\midrule
\textbf{Input} & 15 observed boxes $[c_x, c_y, w, h]$, normalized & The 15 observed boxes rendered as text, together with the observation frames \\
\addlinespace
\textbf{Output} & 45 future boxes; stochastic models emit $K$ samples & 45 future boxes parsed from text  \\
\bottomrule
\end{tabular}
\end{table}

\noindent\textbf{Metrics.} We evaluate trajectory quality with displacement and overlap metrics{, following the protocol established for bounding-box forecasting in the image plane~\cite{citywalks}}. The Average Displacement Error (ADE) and Final Displacement Error (FDE) measure the $\ell_2$ distance between predicted and ground-truth box centers, averaged over the
$T=45$ predicted steps and at the final step,
\begin{equation}
\mathrm{ADE}=\frac{1}{T}\sum_{t=1}^{T}\|\hat{p}_t-p_t\|_2, \quad
\mathrm{FDE}=\|\hat{p}_T-p_T\|_2,
\end{equation}
where $p_t$ and $\hat{p}_t$ are the ground-truth and predicted centers at step $t$. As agents appear at widely varying scales under the ego-perspective, raw pixel errors favor distant, small-box agents{, an effect previously
noted for egocentric prediction~\cite{rasouli_scale}}. We therefore report the scale-normalized variants sADE and sFDE, {which we introduce here and which normalize} each displacement by the ground-truth box diagonal $d_t$,
\begin{equation}
\mathrm{sADE}=\frac{1}{T}\sum_{t=1}^{T}\frac{\|\hat{p}_t-p_t\|_2}{d_t}, \quad
\mathrm{sFDE}=\frac{\|\hat{p}_T-p_T\|_2}{d_T}.
\end{equation}
Finally, the Average and Final Bounding-box IoU (AIoU/FIoU) measure box overlap, averaged over the horizon and at the final step,
\begin{equation}
\mathrm{AIoU}=\frac{1}{T}\sum_{t=1}^{T}\mathrm{IoU}(\hat{b}_t,b_t), \quad
\mathrm{FIoU}=\mathrm{IoU}(\hat{b}_T,b_T).
\end{equation}
{ADE, FDE, AIoU, and FIoU follow the definitions used
in~\cite{citywalks}, keeping our results directly comparable with existing image-plane forecasting benchmarks.} Lower values are better for the displacement metrics (ADE/FDE, sADE/sFDE), and higher values are better for the overlap metrics (AIoU/FIoU). For stochastic models, the single reported trajectory is the mean over their $K$ samples; deterministic and VLM models output a single trajectory directly.

\section{Additional Studies and Analysis}

\subsection{Exploratory Study: Parameter-Efficient Adaptation of VLMs}
\label{sec:supp_lora}

Sec.~4.5 of the main paper reports a preliminary study in which two VLMs are
adapted with Low-Rank Adaptation (LoRA)~\cite{lora}: Qwen3-VL-2B~\cite{qwen}
and UniDriveVLA~\cite{unidrivevla}. UniDriveVLA is built on
Qwen3-VL-2B-Instruct with the same architecture and chat template, so the two
models differ mainly in driving pretraining. This section gives the training
setup and the full results.

\noindent\textbf{Experimental Setup.}
LoRA adapters are applied to the query, key, value, and output projection matrices of the language model with rank $r=16$, $\alpha=32$, and dropout $0.05$. For both models, only 6.42M parameters ($0.30\%$ of the 2.13B model parameters) are updated. Models are trained for three epochs using AdamW in bf16 precision with a learning rate of $1\times10^{-4}$, cosine learning-rate scheduling, a warmup ratio of $0.03$, an effective batch size of 8, and gradient checkpointing. The loss is computed on the answer tokens only, and the adapter from the last epoch is used. The visual inputs, prompts, and output formats are identical to those used in the zero-shot evaluation (Tables~\ref{tab:supp_erma_io} and~\ref{tab:supp_tatp_io}). For ERMA, a separate adapter is trained for each $\tau$. For TATP, adapters are trained on one full-horizon window per training track (2,544 windows), whereas the conventional models use all 21,805 training windows.

\noindent\textbf{Results.}
Table~\ref{tab:supp_lora} reports all metrics; the findings are discussed in Sec.~4.5 of the main paper. Both adapted models parse 100\% of their answers on both tasks, and every TATP answer contains all 45 boxes.

\begin{table}[h]
\centering
\caption{\textbf{Zero-shot \vs LoRA-adapted VLMs.} ERMA reports Acc / weighted-F1 / macro-F1 (\%) on the test set; TATP is evaluated on the 1,246-window subset used in the main paper. The best conventional model is shown for reference.}
\label{tab:supp_lora}
\scriptsize
\setlength{\tabcolsep}{3pt}
\begin{tabular*}{\linewidth}{@{\extracolsep{\fill}} l cccc}
\toprule
\multicolumn{5}{l}{\textbf{ERMA} (Acc / weighted-F1 / macro-F1)} \\
\midrule
\textbf{Model} & $\tau{=}0.2$ & $\tau{=}0.3$ & $\tau{=}0.5$ & $\tau{=}1.0$ \\
\midrule
Qwen3-VL-2B & 28.5/15.7/10.6 & 28.9/16.1/10.5 & 30.0/16.6/10.8 & 29.3/17.1/11.6 \\
\quad + LoRA & 46.3/44.5/36.0 & 47.0/45.5/36.4 & 67.8/64.9/46.6 & 80.7/78.2/56.5 \\
UniDriveVLA & 22.2/16.0/12.8 & 24.8/19.0/16.1 & 33.7/30.1/24.6 & 39.6/36.6/23.0 \\
\quad + LoRA & 50.4/49.1/41.3 & 55.2/54.0/44.9 & 76.3/74.5/58.5 & 85.9/84.4/67.3 \\
\midrule
Best backbone & 51.1/49.7/43.4 & 50.0/46.9/38.1 & 53.3/50.5/38.7 & 58.5/53.8/43.2 \\
 & (SlowFast) & (SlowFast) & (I3D) & (X3D) \\
\bottomrule
\end{tabular*}

\vspace{6pt}

\begin{tabular*}{\linewidth}{@{\extracolsep{\fill}} l cccccc}
\toprule
\multicolumn{7}{l}{\textbf{TATP}} \\
\midrule
\textbf{Model} & ADE\,$\downarrow$ & FDE\,$\downarrow$ & sADE\,$\downarrow$ & sFDE\,$\downarrow$ & AIoU\,$\uparrow$ & FIoU\,$\uparrow$ \\
\midrule
Qwen3-VL-2B & 91.7 & 187.2 & 0.690 & 1.247 & 0.370 & 0.156 \\
\quad + LoRA & 58.7 & 140.8 & 0.439 & 0.945 & 0.515 & 0.242 \\
UniDriveVLA & 95.1 & 192.3 & 0.693 & 1.229 & 0.354 & 0.147 \\
\quad + LoRA & 59.6 & 142.9 & 0.448 & 0.969 & 0.511 & 0.238 \\
\midrule
SGNet & 51.8 & 115.0 & 0.389 & 0.785 & 0.549 & 0.312 \\
\bottomrule
\end{tabular*}
\end{table}

\subsection{Full-Test-Set TATP Results}
\label{sec:supp_tatp_full}

Table~\ref{tab:supp_tatp_full} reports the conventional methods on all 10,133
test windows, together with ADE at three prediction horizons. SGNet remains the
best method on every metric, and errors grow steadily with the horizon for all
methods. On the 1,246-window subset used in the main paper, every method keeps
its rank on ADE, FDE, sADE and AIoU, and its ADE is about 10\% higher.

\begin{table}[h]
\centering
\caption{\textbf{TATP results of the conventional methods on all 10,133 test
windows.} Bottom: ADE per prediction horizon, and full-set \vs subset ADE.
\textbf{Best} and \underline{second-best} per column in bold and underline.}    
\label{tab:supp_tatp_full}
\scriptsize
\setlength{\tabcolsep}{3pt}
\begin{tabular*}{\linewidth}{@{\extracolsep{\fill}} l cccccc}
\toprule
\textbf{Model} & ADE\,$\downarrow$ & FDE\,$\downarrow$ & sADE\,$\downarrow$ & sFDE\,$\downarrow$ & AIoU\,$\uparrow$ & FIoU\,$\uparrow$ \\
\midrule
PIETraj~\cite{PIETraj} & 66.0 & 125.0 & 0.439 & \underline{0.755} & 0.474 & \underline{0.307} \\
BiTraP~\cite{bitrap} & \underline{51.0} & \underline{120.0} & \underline{0.347} & 0.805 & \underline{0.586} & 0.305 \\
ABC+~\cite{ABC+} & 53.2 & 126.3 & 0.385 & 0.901 & 0.578 & 0.296 \\
SGNet~\cite{SGNet} & \textbf{46.9} & \textbf{103.8} & \textbf{0.309} & \textbf{0.648} & \textbf{0.614} & \textbf{0.376} \\
\bottomrule
\end{tabular*}

\vspace{6pt}

\begin{tabular*}{\linewidth}{@{\extracolsep{\fill}} l ccc ccc}
\toprule
 & \multicolumn{3}{c}{ADE by horizon\,$\downarrow$} & \multicolumn{3}{c}{ADE: full \vs subset} \\
\cmidrule(lr){2-4} \cmidrule(lr){5-7}
\textbf{Model} & 0.5\,s & 1.0\,s & 1.5\,s & Full & Subset & Ratio \\
\midrule
PIETraj~\cite{PIETraj} & 40.4 & 52.2 & 66.0 & 66.0 & 73.3 & 1.11 \\
BiTraP~\cite{bitrap} & \textbf{17.5} & \underline{33.1} & \underline{51.0} & 51.0 & 55.5 & 1.09 \\
ABC+~\cite{ABC+} & \underline{17.8} & 33.9 & 53.2 & 53.2 & 58.5 & 1.10 \\
SGNet~\cite{SGNet} & \textbf{17.5} & \textbf{31.2} & \textbf{46.9} & 46.9 & 51.8 & 1.10 \\
\bottomrule
\end{tabular*}
\end{table}

\subsection{ERMA Confusion Matrix Analysis}

\begin{figure*}[p]
    \centering
    \includegraphics[width=\textwidth]{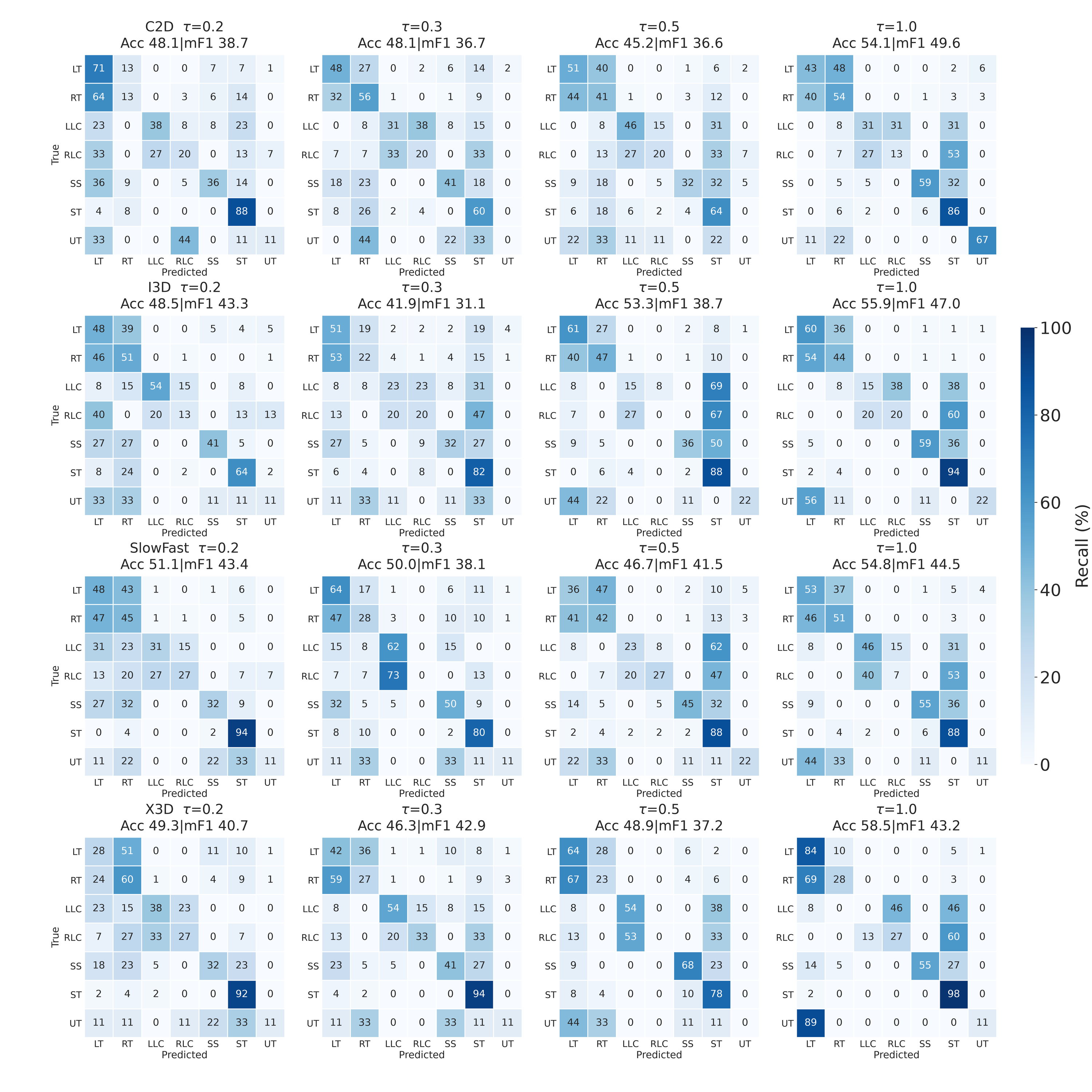}
    \caption{\textbf{Row-normalized confusion matrices (\%) for all four video
    backbones across observation ratios $\tau$.} Rows correspond to C2D, I3D,
    SlowFast, and X3D; columns to $\tau \in \{0.2, 0.3, 0.5, 1.0\}$. Each panel is
    titled with its Accuracy and macro-F1.}
    \label{fig:supp_backbone_confusion}
\end{figure*}

\noindent\textbf{Conventional Models.} Fig.~\ref{fig:supp_backbone_confusion} presents the row-normalized confusion matrices of the four conventional backbones across all observation ratios, complementing the analysis in the main paper. A consistent error pattern emerges across all architectures. Straight (ST) has the highest recall in every panel, between 60\% and 98\%. Confusion between opposite turning directions (LT$\leftrightarrow$RT) persists at every observation ratio. Even at $\tau=1.0$, 40--69\% of right turns are predicted as left turns, so turn direction is not resolved by observing the completed maneuver. The lane-change classes are predicted mostly as Straight or as the opposite lane change, even with full observation. Right Lane Change recall never exceeds 33\%, making the lane changes a principal contributor to the lower macro-F1 scores. Slow/Stop (SS) is frequently misclassified as a turn at $\tau=0.2$ (41--59\% of clips) and increasingly as Straight with longer observation. Finally, U-Turn (UT) remains the hardest class. Its recall stays at or below 22\%, except for C2D at $\tau=1.0$ (67\%), and at full observation it is often predicted as Left Turn (up to 89\%).

\begin{figure*}[p]
    \centering
    \includegraphics[width=\textwidth]{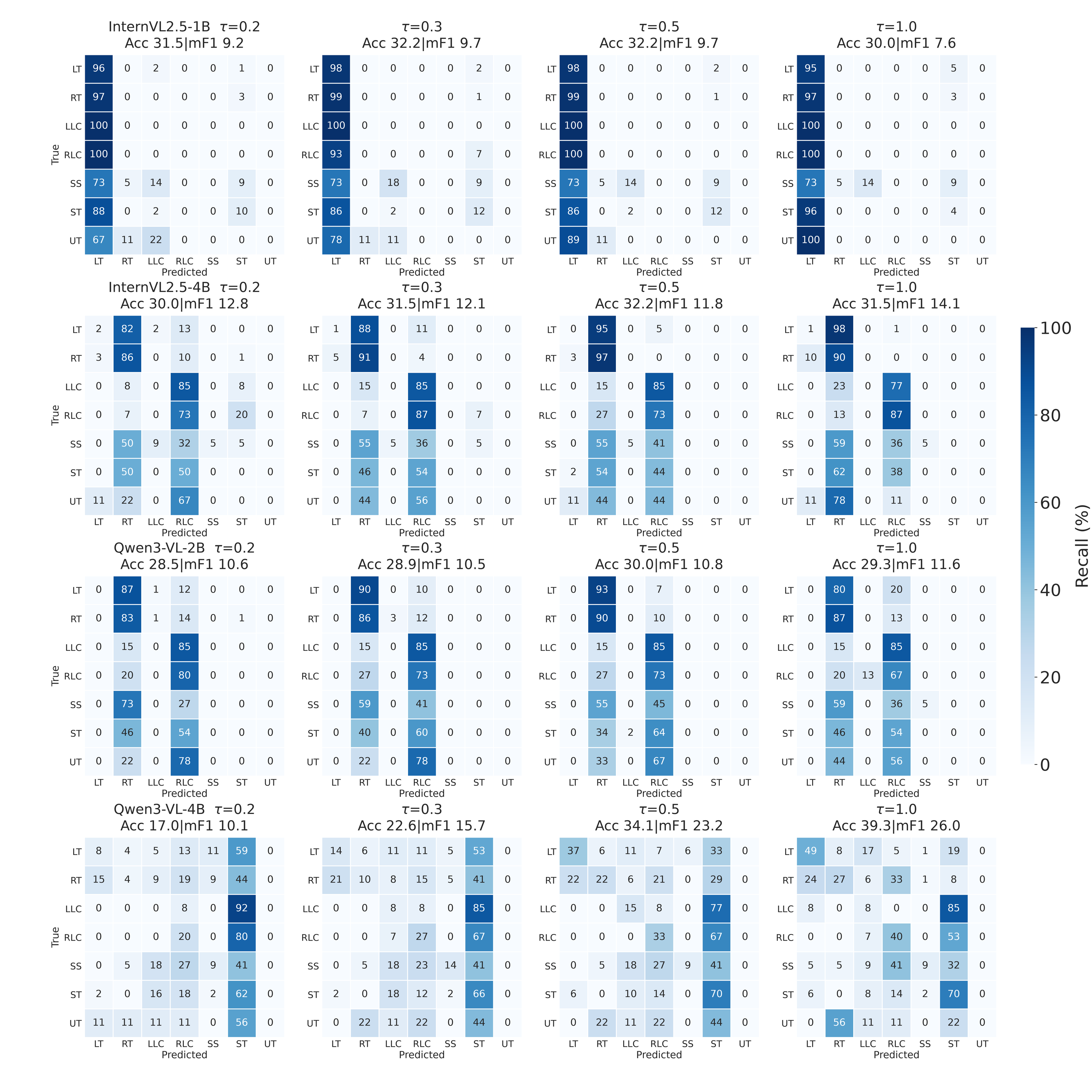}
    \caption{\textbf{Row-normalized confusion matrices (\%) for all four zero-shot
    VLMs across observation ratios $\tau$.} Rows correspond to InternVL2.5-1B/4B
    and Qwen3-VL-2B/4B; columns to $\tau \in \{0.2, 0.3, 0.5, 1.0\}$. Three of the
    four models collapse onto one or two classes at every observation ratio,
    visible as solid predicted columns.}
    \label{fig:supp_vlm_confusion}
\end{figure*}

\noindent\textbf{Vision-Language Models.} Fig.~\ref{fig:supp_vlm_confusion} reveals a markedly different failure mode for zero-shot VLMs. Rather than exhibiting structured inter-class confusion, three of the four models collapse onto one or two classes at every observation ratio. InternVL2.5-1B predicts Left Turn (LT) for most clips (67--100\% of each class). InternVL2.5-4B and Qwen3-VL-2B predict Right Turn (RT) or Right Lane Change (RLC) for over 80\% of the clips in every class, and Qwen3-VL-2B never predicts Left Turn. Qwen3-VL-4B is the exception. It predicts mostly Straight (ST) at early observation ratios but diversifies with longer observation, raising its Left Turn recall from 8\% to 49\%. It never predicts U-Turn, and its discriminative capability remains substantially below that of the fine-tuned conventional models.

To better understand this behavior, we manually inspected the outputs of all {four general-purpose} zero-shot VLMs {on ERMA}. Contrary to our initial hypothesis, the poor performance was not caused by malformed responses, prompt-following failures, or parsing errors. Instead, every model consistently produced valid outputs by selecting one of the seven
maneuver classes exactly as instructed. The observed degradation therefore stems from a systematic prediction bias, suggesting that current general-purpose VLMs struggle to infer fine-grained rider intentions from short ego-centric MTW videos without domain-specific adaptation.

\begin{figure*}[p]
    \centering
    \includegraphics[width=\textwidth]{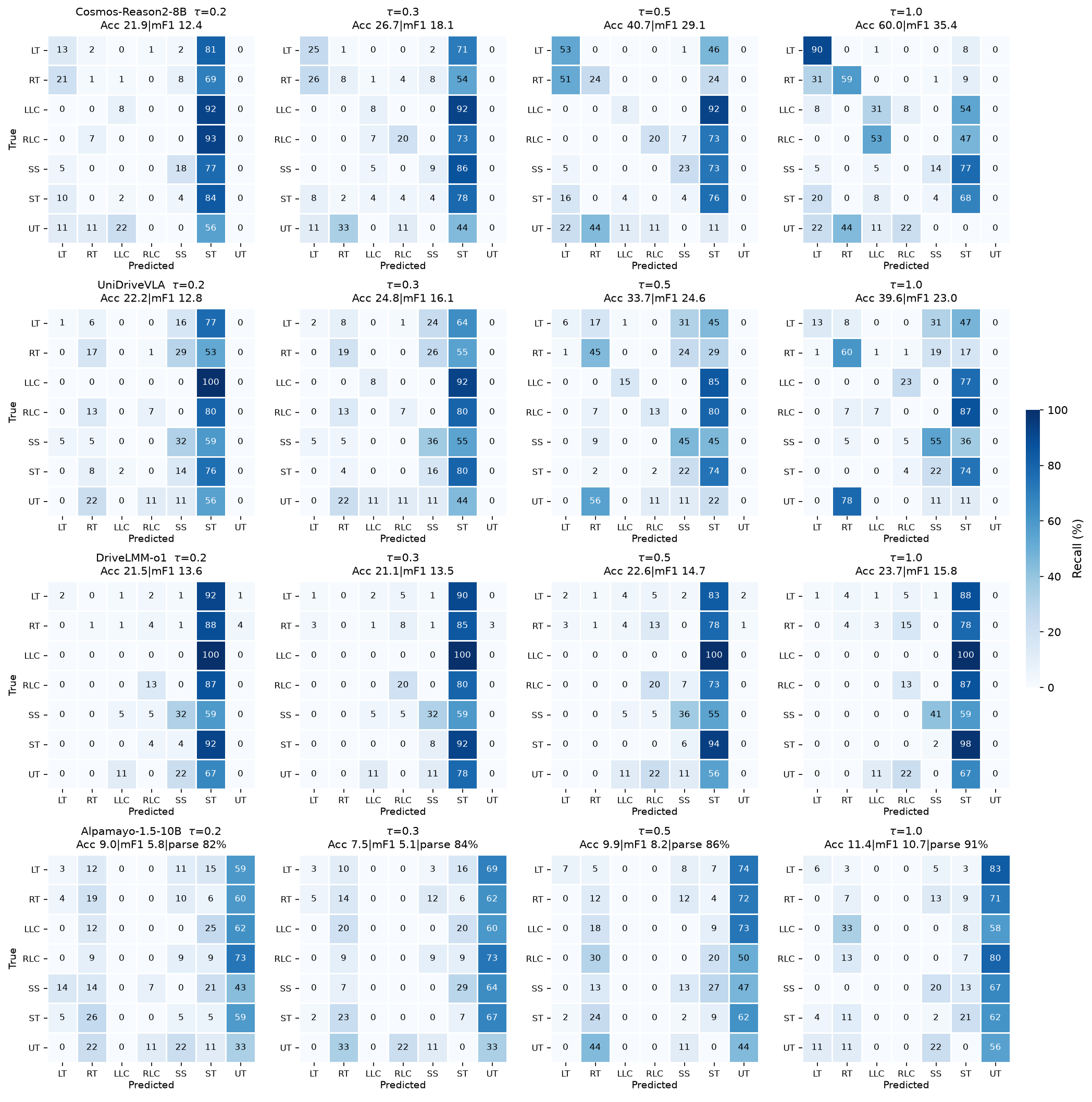}
    \caption{\textbf{Row-normalized confusion matrices (\%) for the four
    driving-specific VLMs and VLAs across observation ratios $\tau$.} Rows
    correspond to Cosmos-Reason2-8B, UniDriveVLA, DriveLMM-o1, and
    Alpamayo-1.5-10B; columns to $\tau \in \{0.2,0.3,0.5,1.0\}$. Cosmos-Reason2-8B
    develops the clearest diagonal, while DriveLMM-o1 and Alpamayo-1.5-10B
    collapse onto Straight and U-Turn respectively.}
    \label{fig:driving_confusion}
\end{figure*}

\noindent\textbf{Driving-specific VLMs and VLAs.} Fig.~\ref{fig:driving_confusion} shows the corresponding matrices for the four models pretrained on driving data. Cosmos-Reason2-8B develops the clearest diagonal among the zero-shot models, and it does so progressively: left-turn recall rises from 13\% at $\tau=0.2$ to 90\% at $\tau=1.0$, and right-turn recall from 1\% to 59\%. Its residual errors are concentrated on Straight, which absorbs 47--93\% of both lane-change classes at every ratio, and it never predicts U-Turn. UniDriveVLA recognizes Straight (74--80\%) and increasingly Right Turn and Slow/Stop, but rarely Left Turn (at most 13\%); most left turns are predicted as Straight or Slow/Stop. The remaining two models collapse toward a single class throughout. DriveLMM-o1 assigns more than half of every true class to Straight, and Alpamayo-1.5-10B assigns 43--83\% of every class to U-Turn; both collapses persist at all four observation ratios. Neither model therefore distinguishes between maneuver categories, and their accuracies largely reflect the frequency of the predicted class in the test set.

\begin{figure*}[p]
    \centering
    \includegraphics[width=\textwidth]{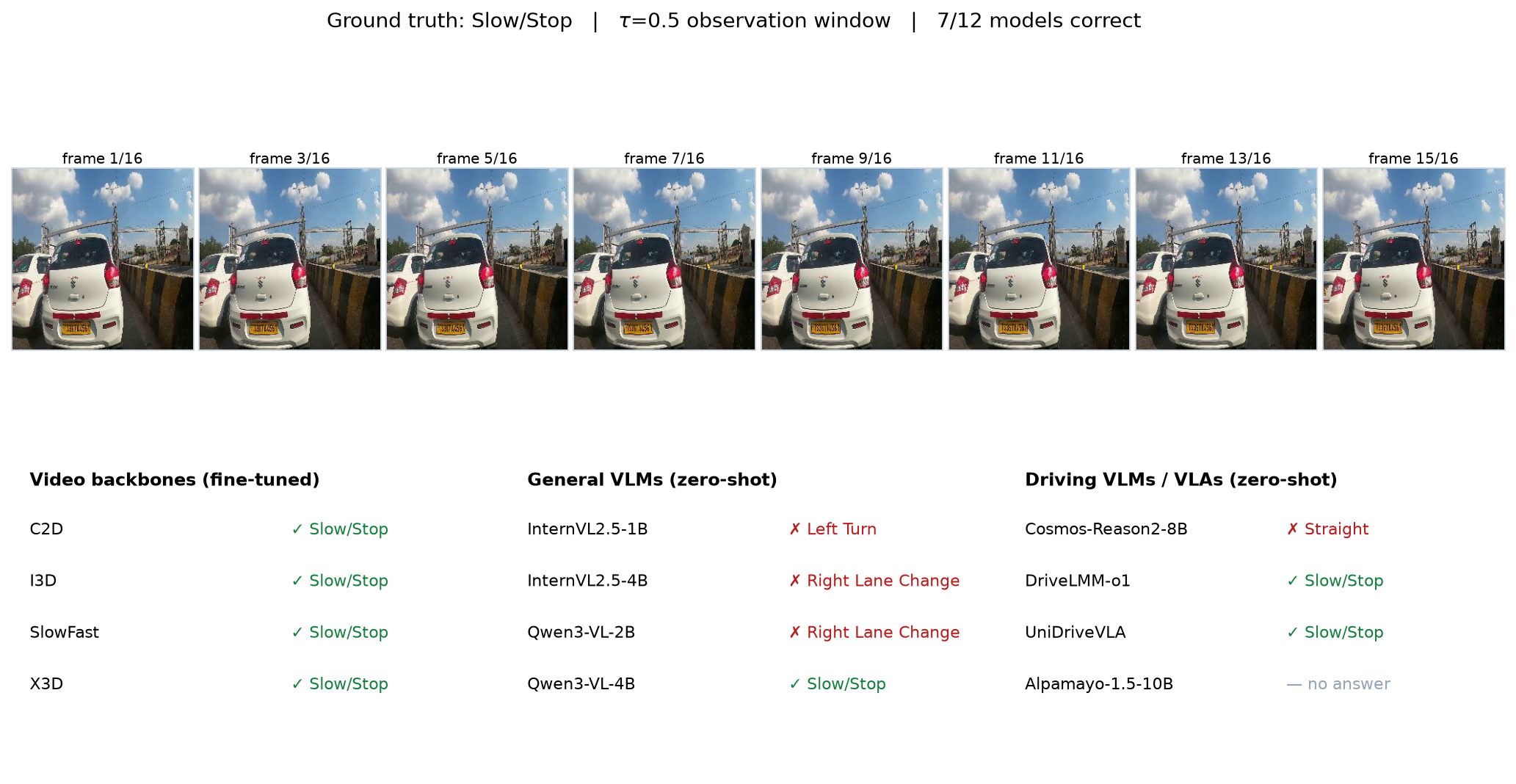}
    \caption{{\textbf{Slow/Stop, dense stopped traffic ($\tau=0.5$).}
    Three of the four general-purpose VLMs fail, despite clear signs of stopped
    traffic.}}
    \label{fig:qual1}
\end{figure*}

\begin{figure*}[p]
    \centering
    \includegraphics[width=\textwidth]{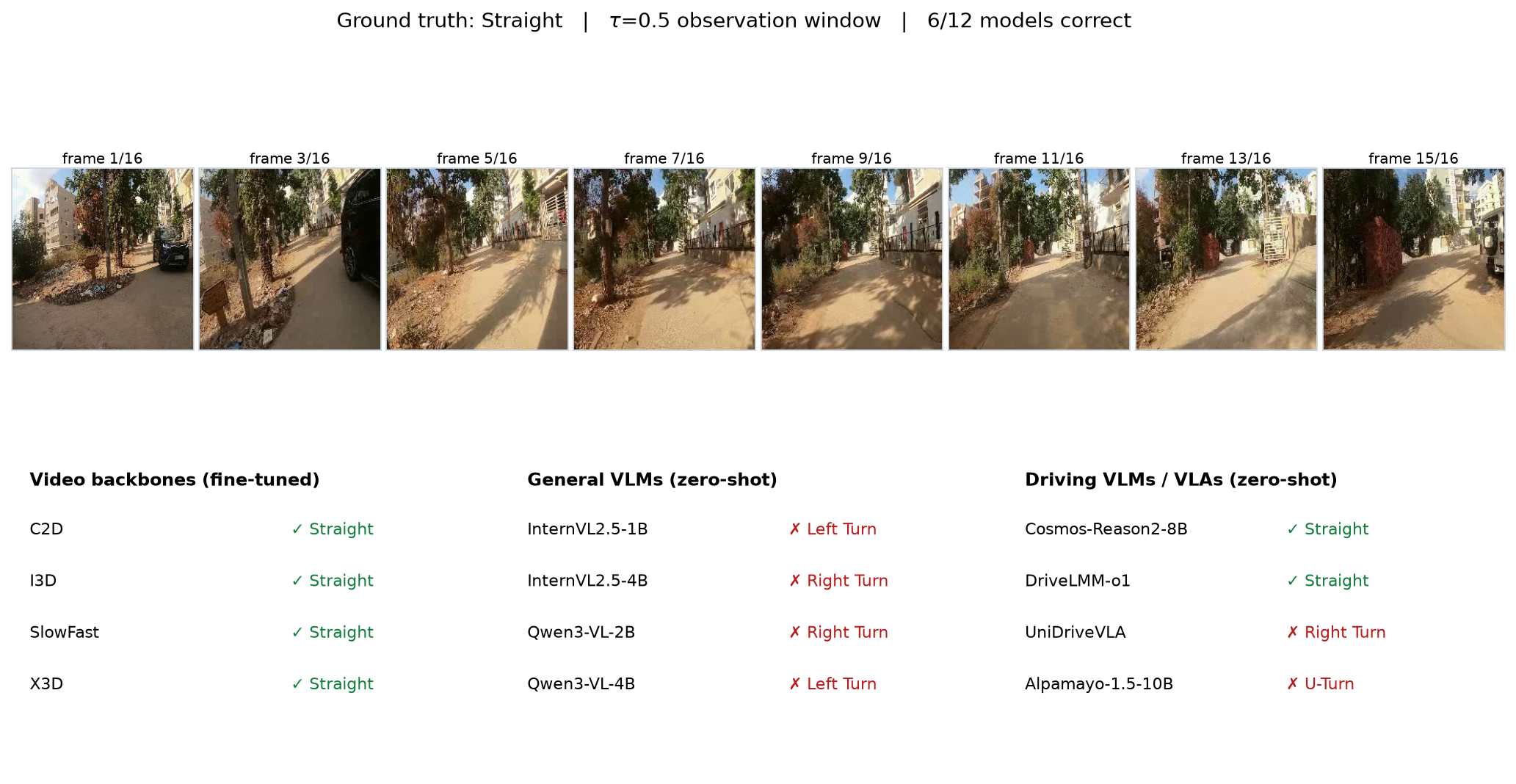}
    \caption{{\textbf{Straight, sparse traffic ($\tau=0.5$).} The
    fine-tuned backbones are correct. All four general-purpose VLMs predict
    unrelated classes.}}
    \label{fig:qual2}
\end{figure*}

\begin{figure*}[p]
    \centering
    \includegraphics[width=\textwidth]{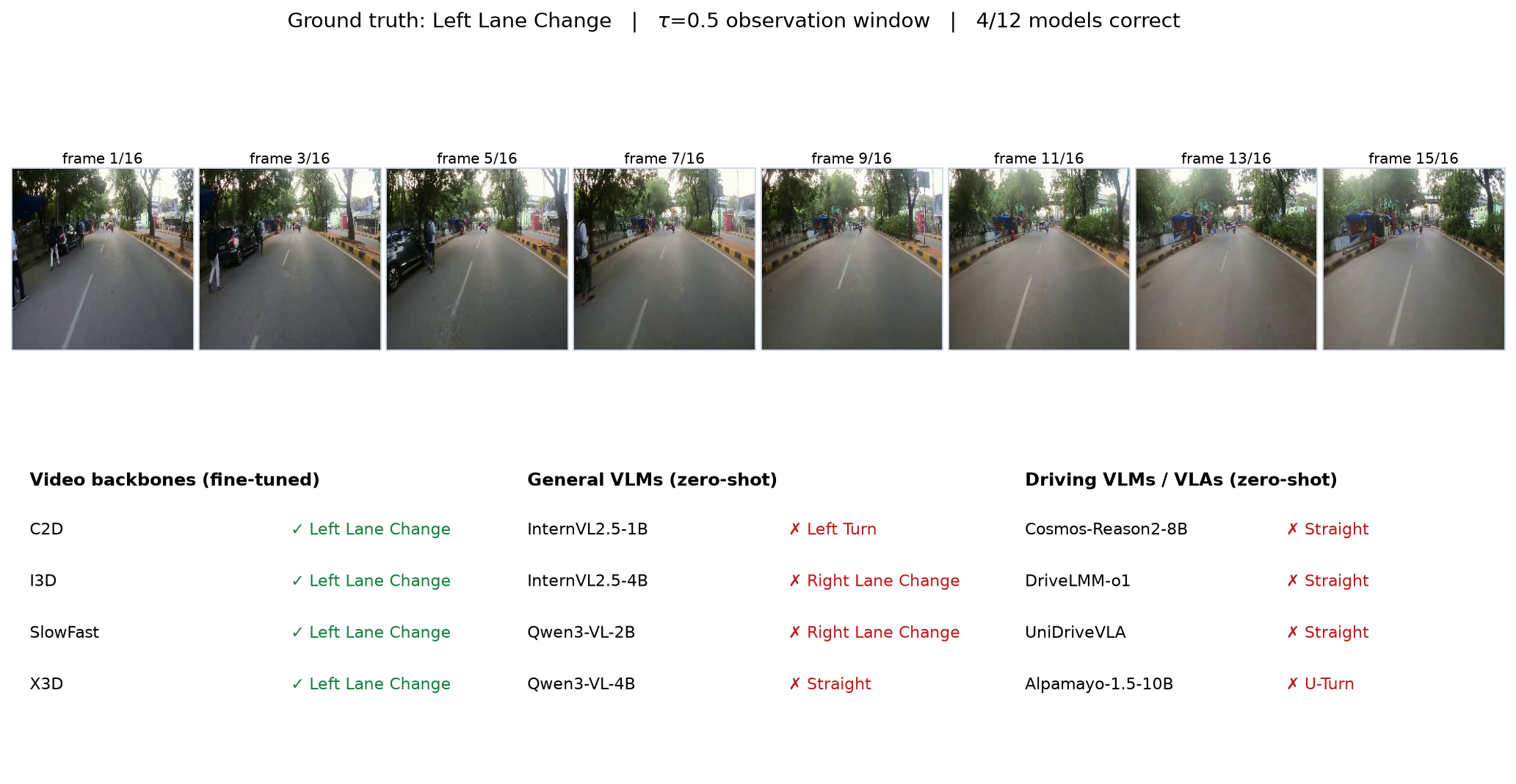}
    \caption{{\textbf{Left Lane Change ($\tau=0.5$).} Only the backbones
    are correct. Most zero-shot models predict Right Lane Change or Straight, as
    in Figs.~\ref{fig:supp_vlm_confusion} and~\ref{fig:driving_confusion}.}}
    \label{fig:qual3}
\end{figure*}

\begin{figure*}[p]
    \centering
    \includegraphics[width=\textwidth]{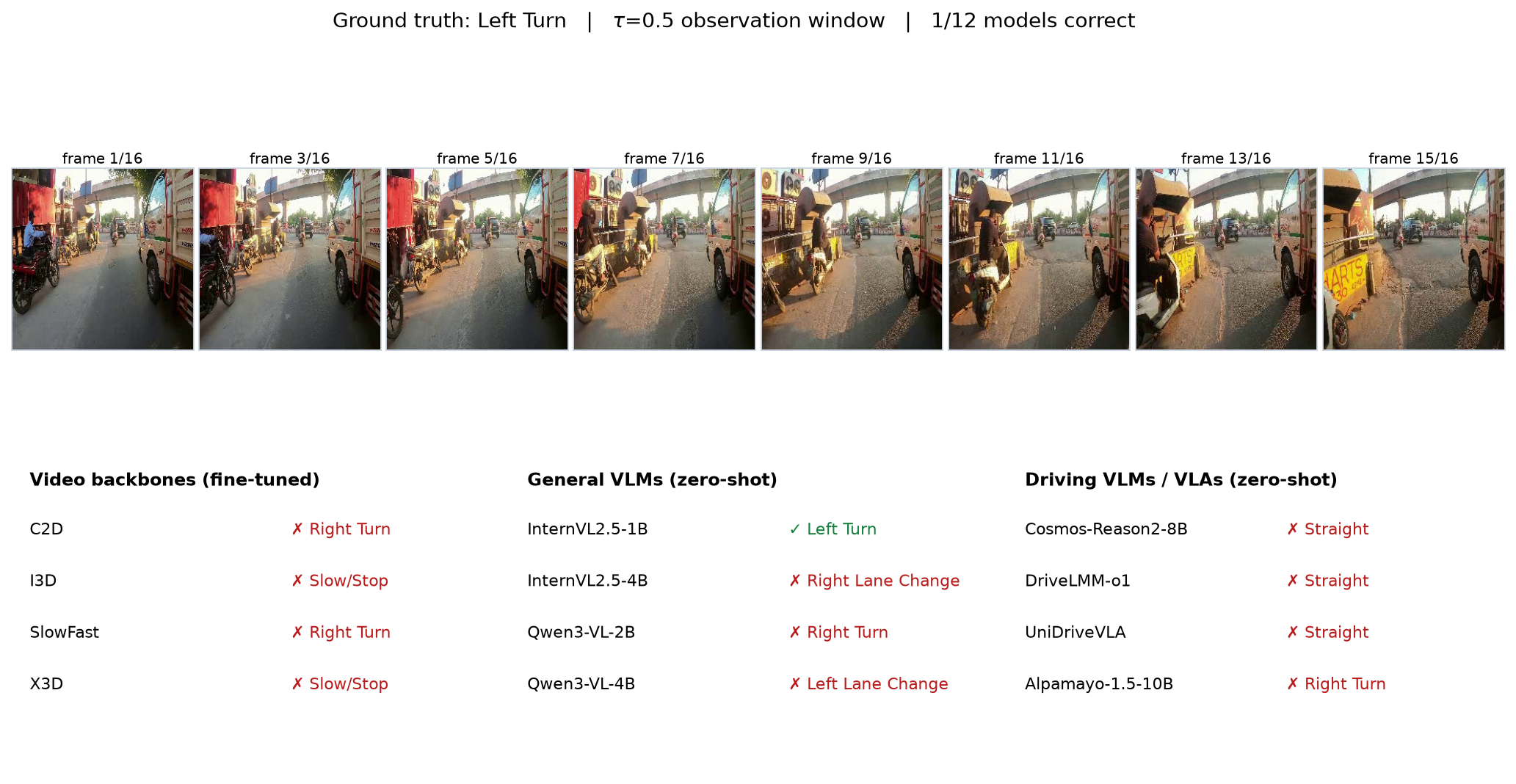}
    \caption{{\textbf{Left Turn, dense unstructured traffic
    ($\tau=0.5$).} Only InternVL2.5-1B is correct, and Left Turn is the class it
    predicts for most clips.}}
    \label{fig:qual4}
\end{figure*}

\begin{figure*}[p]
    \centering
    \includegraphics[width=\textwidth]{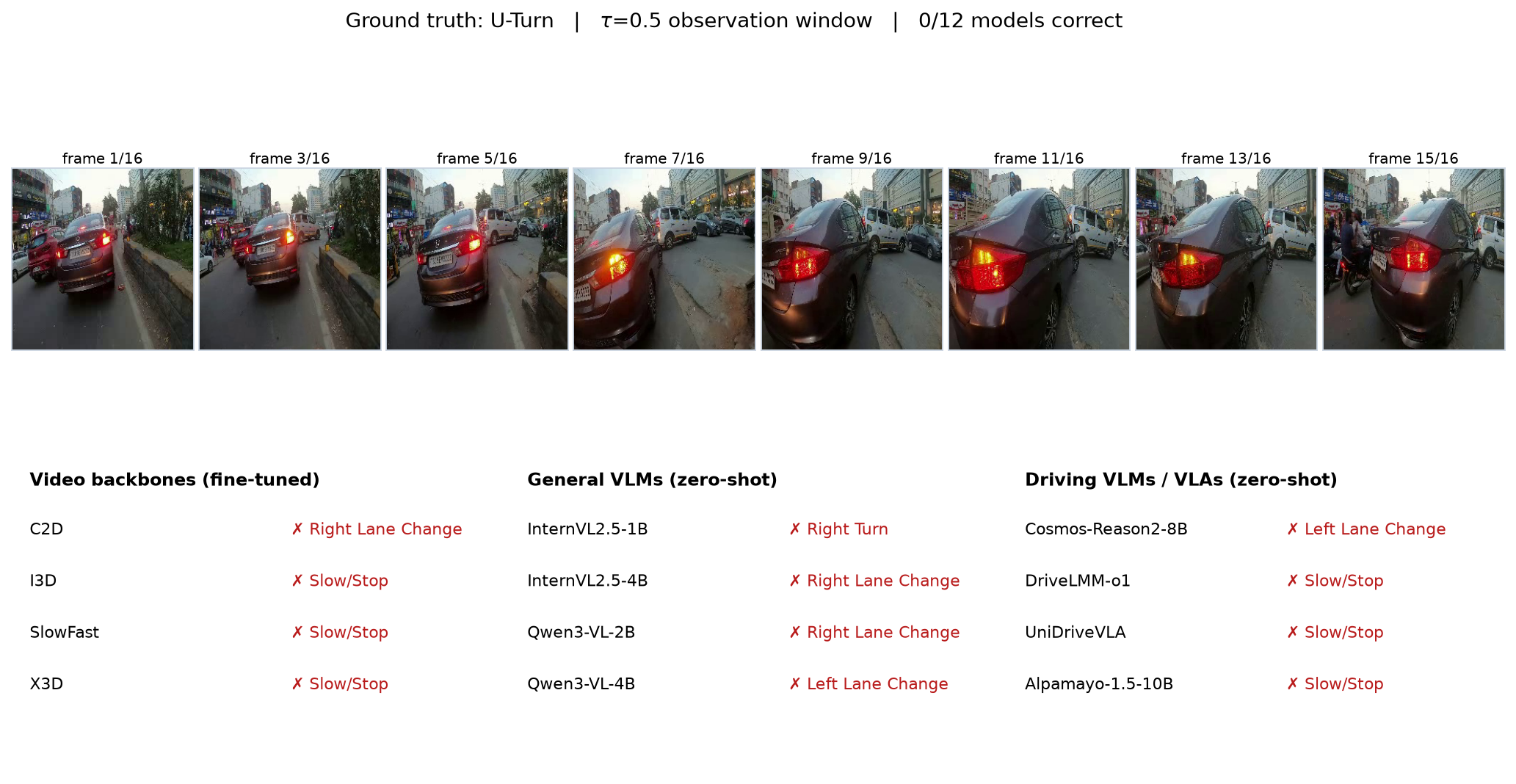}
    \caption{{\textbf{U-Turn ($\tau=0.5$).} No model predicts the correct
    class. U-Turn is the hardest class for every model family.}}
    \label{fig:qual5}
\end{figure*}

\begin{figure*}[p]
    \centering
    \includegraphics[width=\textwidth]{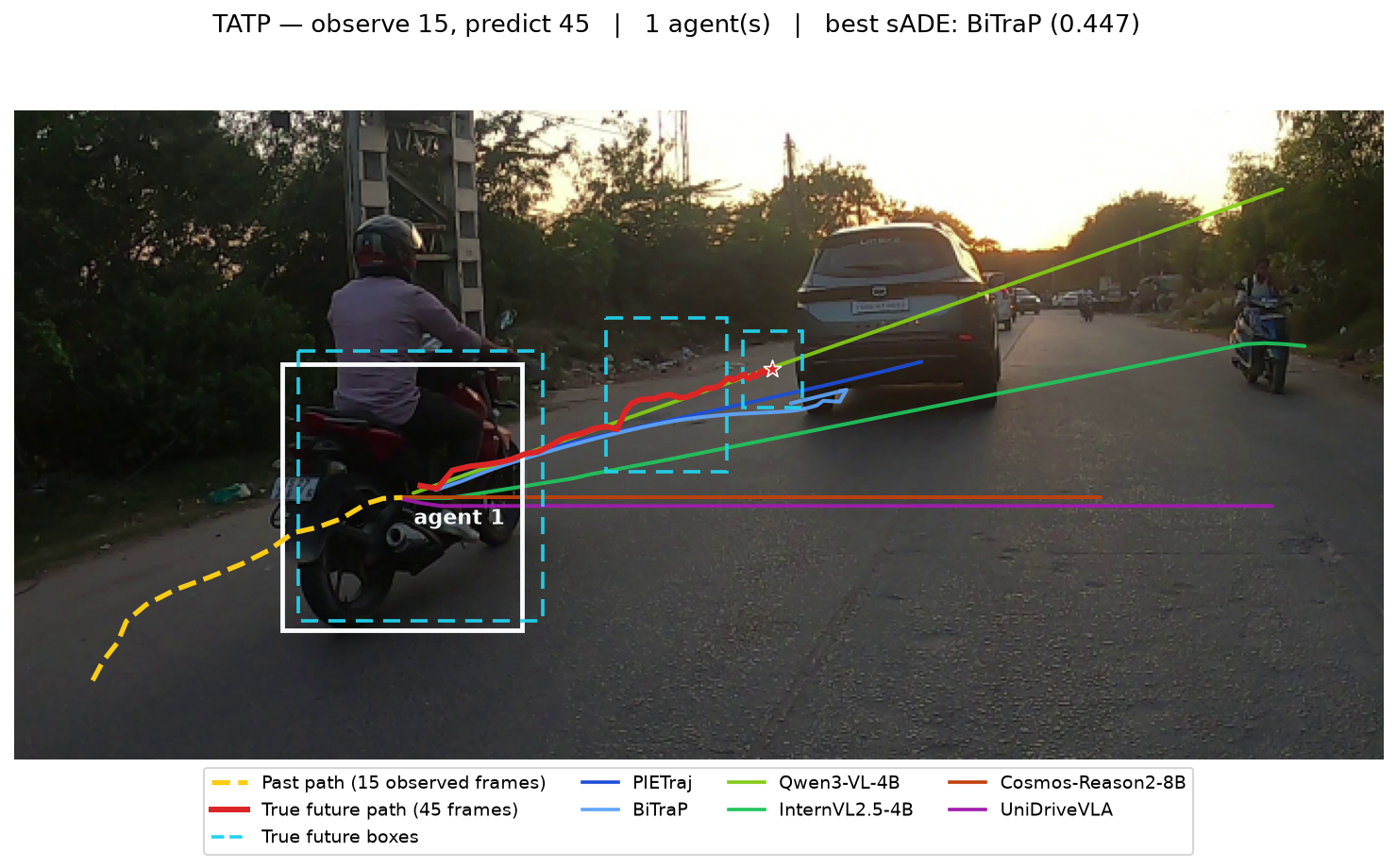}
    \caption{{\textbf{TATP: closely-following MTW.} The fine-tuned
    baselines stay closest to the true path. Several zero-shot predictions
    overshoot it or follow a flat line.}}
    \label{fig:qual6}
\end{figure*}

\begin{figure*}[p]
    \centering
    \includegraphics[width=\textwidth]{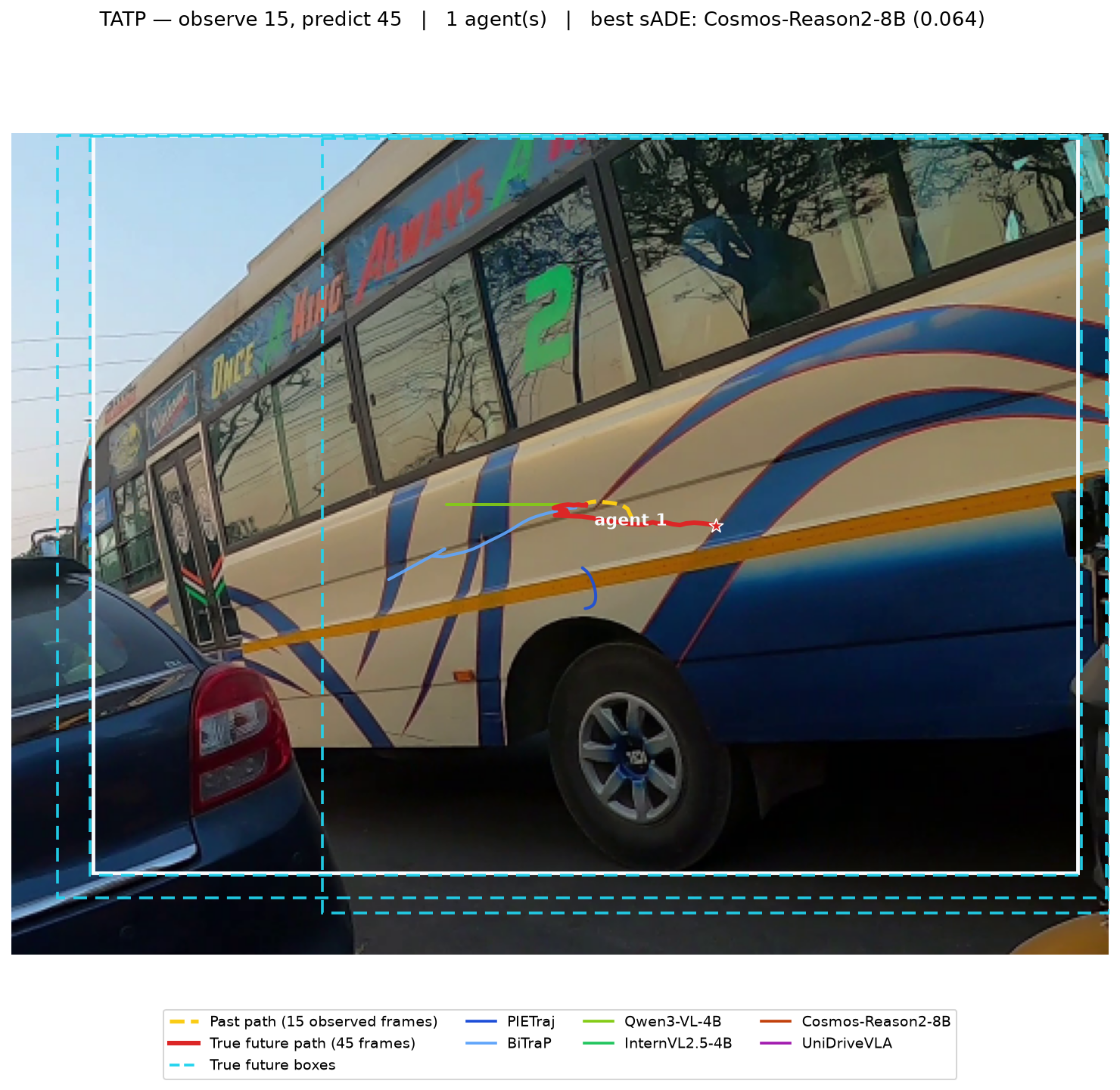}
    \caption{{\textbf{TATP: bus at close range.} The bus barely moves.
    Cosmos-Reason2-8B has the lowest error, while the fine-tuned baselines
    predict motion in the wrong direction.}}
    \label{fig:qual7}
\end{figure*}

\begin{figure*}[p]
    \centering
    \includegraphics[width=\textwidth]{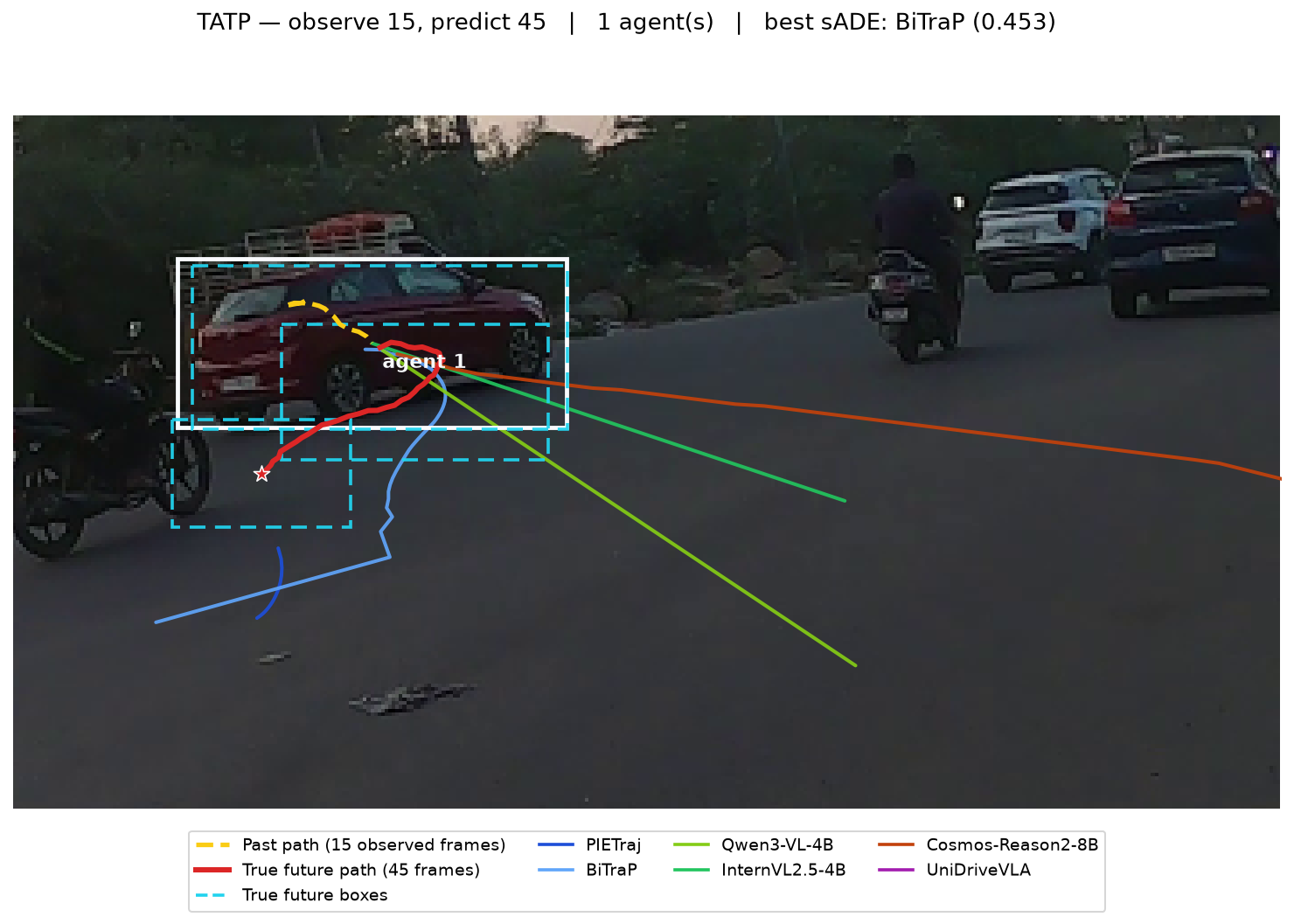}
    \caption{{\textbf{TATP: receding car.} Averaging sampled futures
    gives BiTraP an implausible path. The zero-shot models predict motion in the
    wrong direction.}}
    \label{fig:qual8}
\end{figure*}

{\subsection{Qualitative Analysis}
\paragraph{ERMA.} Figs.~\ref{fig:qual1}--\ref{fig:qual5} show predictions from all twelve models on five test clips at $\tau=0.5$, ordered by the number of models answering correctly. For readability, each figure displays 8 frames sampled across the observation window; the models received the full 16-frame input specified in Table~\ref{tab:supp_erma_io}. The examples mirror the error patterns in the confusion matrices. On clips with clear ego-motion
cues and sparse surroundings, the fine-tuned backbones agree with the ground truth, while the general-purpose VLMs return the classes they predict for most clips regardless of the scene. On the Left Lane Change example, most zero-shot models predict Right Lane Change or Straight, the two classes into which lane changes most often collapse. On the Left Turn through dense unstructured traffic, only InternVL2.5-1B answers correctly, and its prediction coincides with its bias toward Left Turn rather than reflecting scene interpretation. On the U-Turn example no model is correct, consistent with U-Turn being the hardest class for every model family. Alpamayo-1.5-10B returns no parseable answer on one of the five clips.

\paragraph{TATP.} Figs.~\ref{fig:qual6}--\ref{fig:qual8} show predicted bounding-box trajectories on three test clips, with two models drawn from each family for legibility. Each figure overlays the prediction on a single frame; the models observed the full 15-frame input specified in Table~\ref{tab:supp_tatp_io}. The examples reflect the findings of Sec.~4.3 and~4.4 of the main paper. On the closely following MTW, the fine-tuned baselines stay closest to the true path, while several zero-shot predictions overshoot it or follow a flat line. On the bus, which barely moves, Cosmos-Reason2-8B has the lowest error, while the fine-tuned baselines predict motion in the wrong direction. Averaging sampled futures can also yield paths that do not correspond to plausible motion, as visible in BiTraP's prediction for the receding car.}

\begin{figure*}[p]
    \centering
    \includegraphics[width=\textwidth]{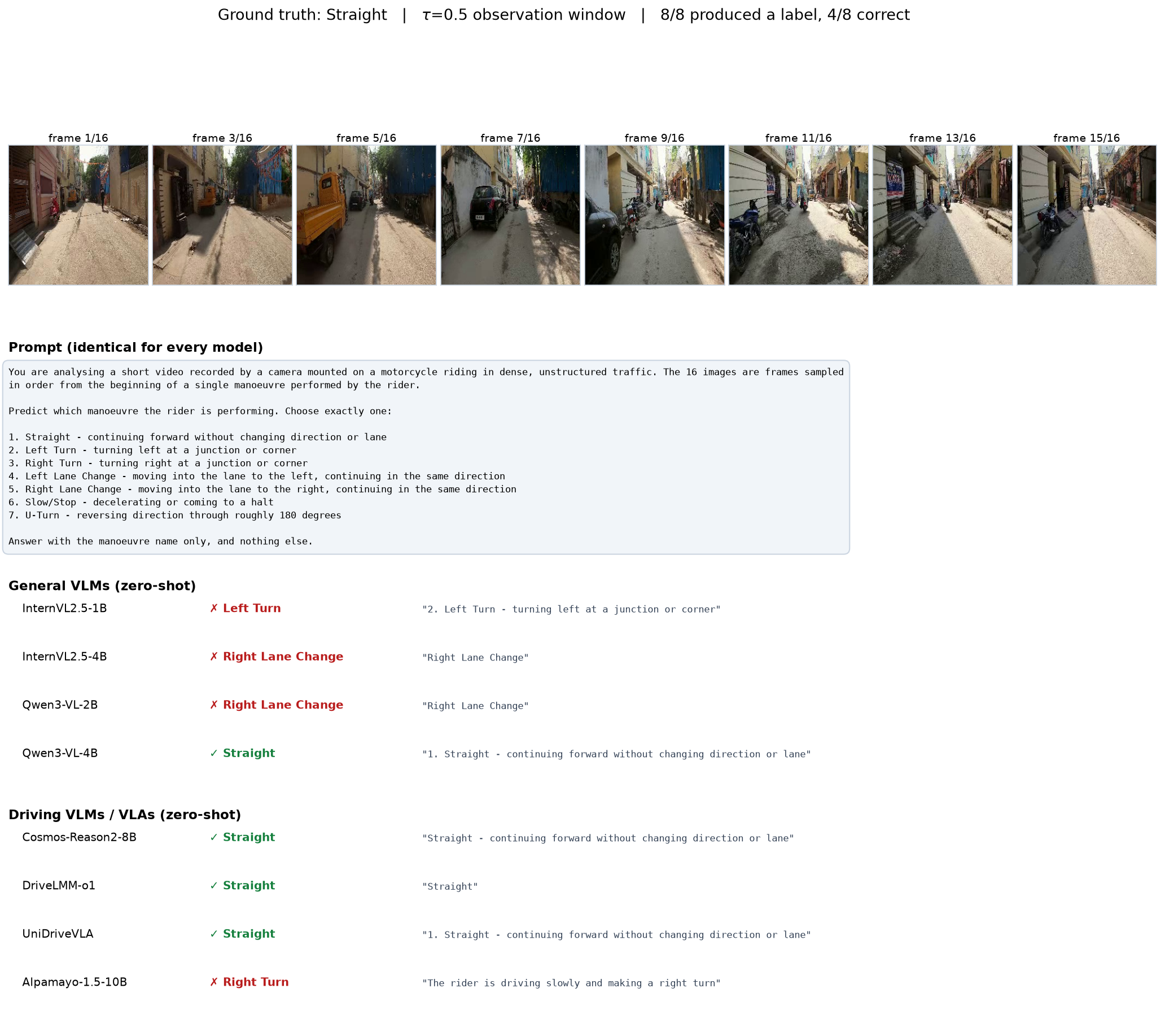}
    \caption{{\textbf{ERMA prompt and responses.} All eight models return
    a parseable class; four are correct.}}
    \label{fig:prompt_erma_ok}
\end{figure*}

\begin{figure*}[p]
    \centering
    \includegraphics[width=\textwidth]{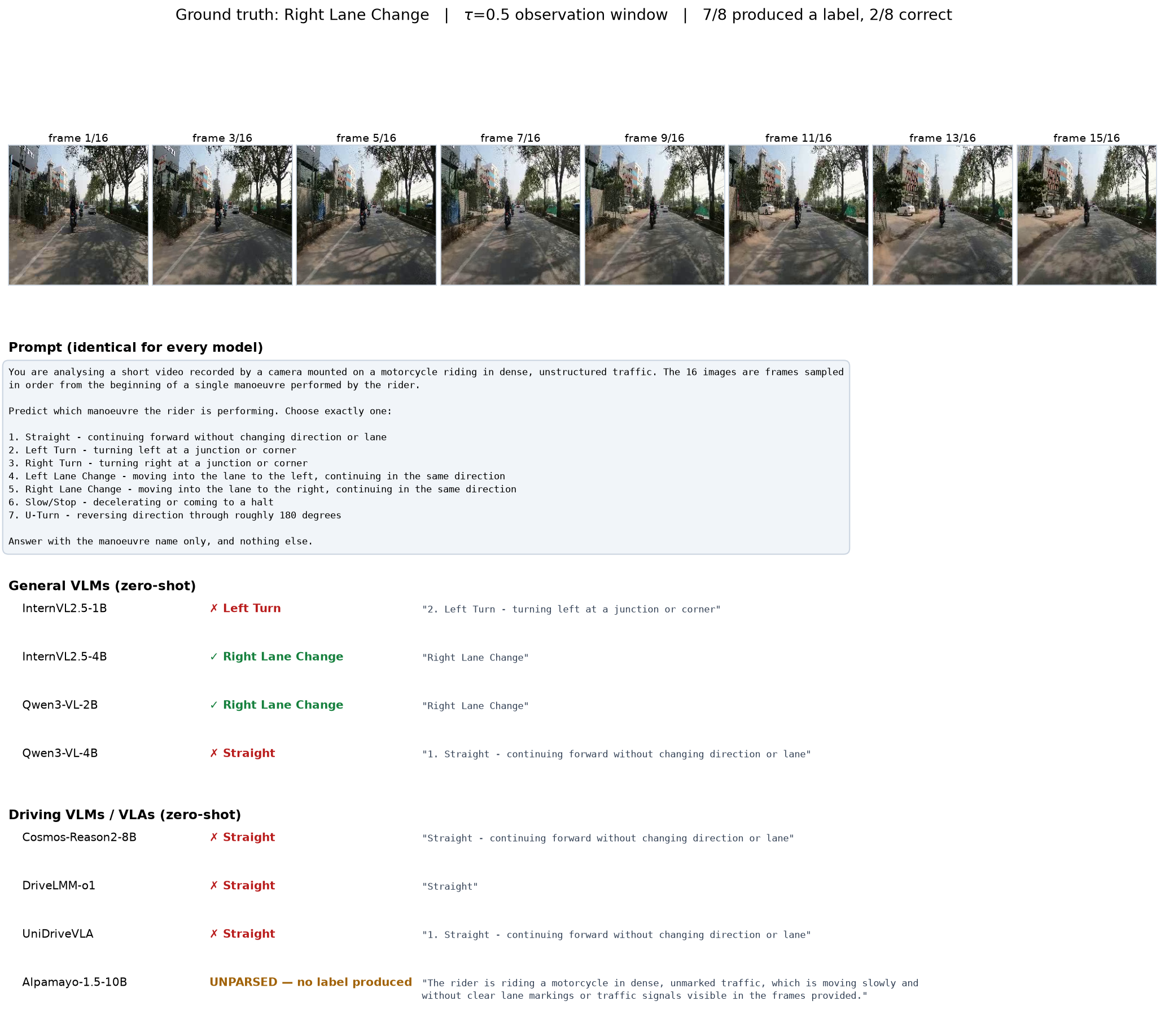}
    \caption{{\textbf{ERMA prompt and responses: unparsed case.}
    Alpamayo-1.5-10B answers in prose, so no class can be extracted.}}
    \label{fig:prompt_erma_fail}
\end{figure*}

\begin{figure*}[p]
    \centering
    \includegraphics[width=\textwidth]{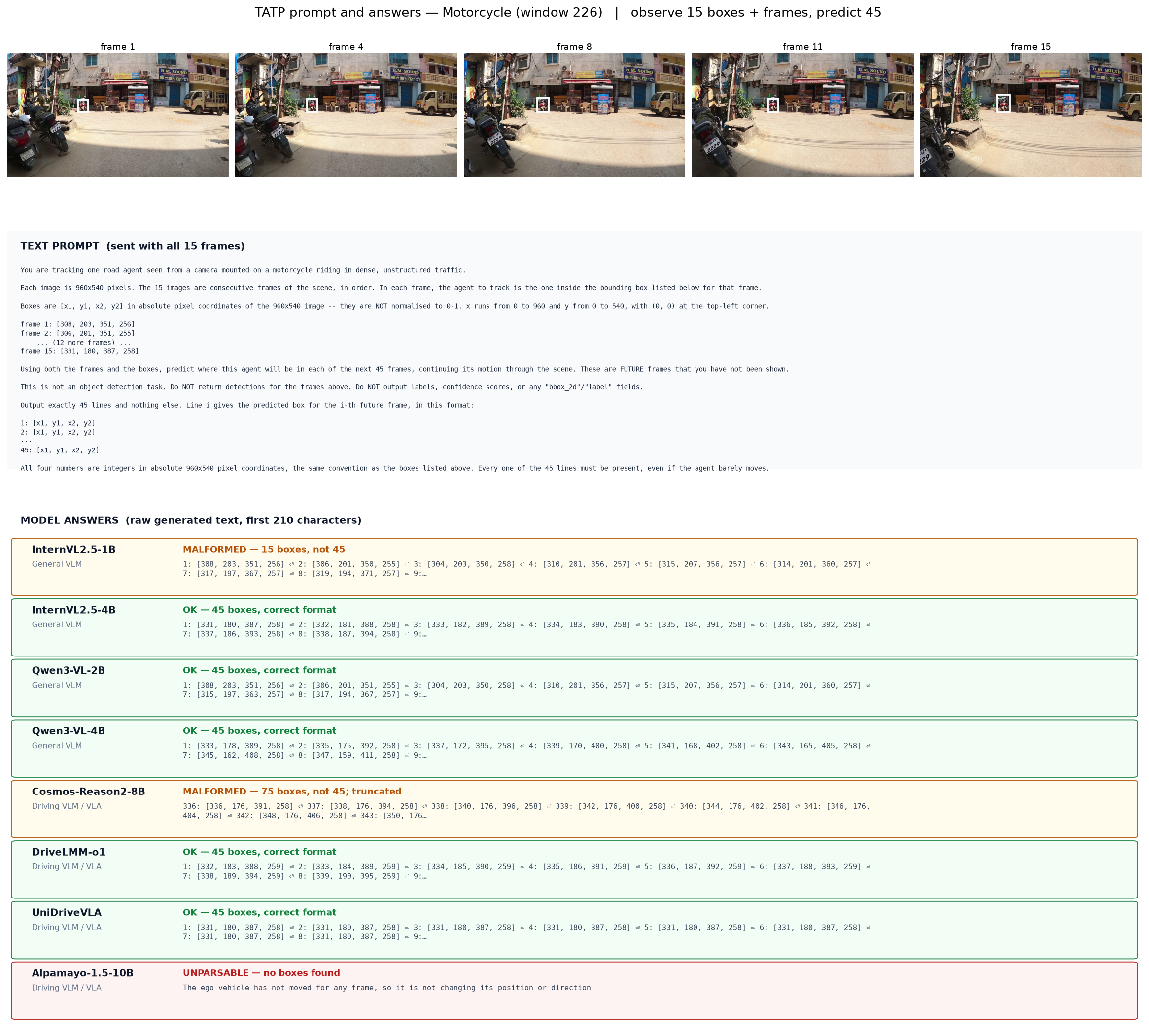}
    \caption{{\textbf{TATP prompt and responses: MTW agent.} Raw output
    of each model. Answers are correctly formatted, malformed or unparsable.}}
    \label{fig:prompt_tatp_mtw}
\end{figure*}

\begin{figure*}[p]
    \centering
    \includegraphics[width=\textwidth]{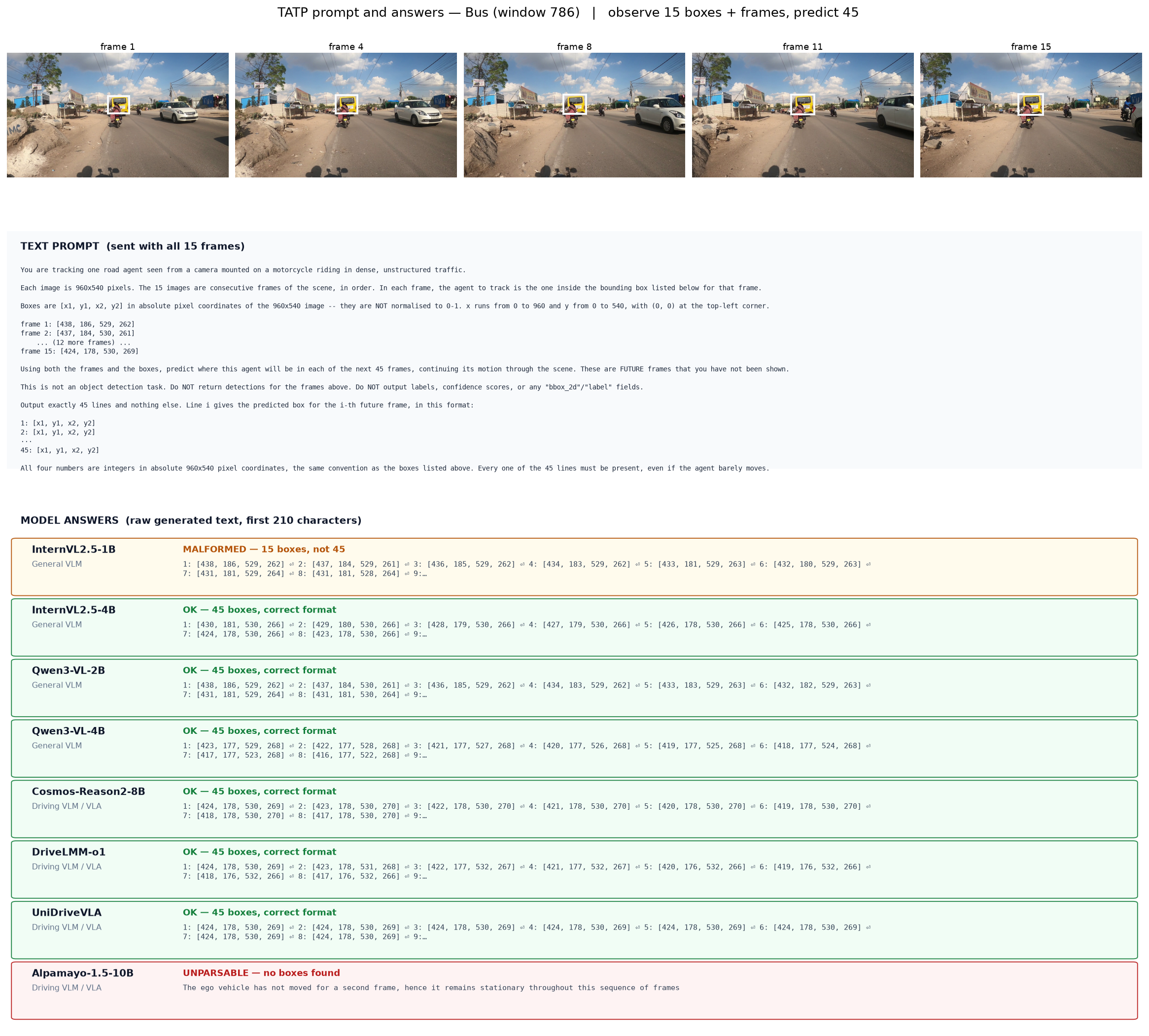}
    \caption{{\textbf{TATP prompt and responses: bus agent.} The same
    three response types occur. InternVL2.5-1B and Qwen3-VL-2B repeat the
    observed boxes as their prediction.}}
    \label{fig:prompt_tatp_bus}
\end{figure*}

{\subsection{Prompts and Model Outputs}
\paragraph{ERMA.} All eight zero-shot models receive an identical prompt listing
the seven maneuver classes with a one-line definition of each and requesting the
class name alone (Figs.~\ref{fig:prompt_erma_ok}--\ref{fig:prompt_erma_fail}).
Most responses comply with this format, so the low accuracies reported in
Sec.~4.2 reflect the choice of class rather than any difficulty in following the
instruction. Alpamayo-1.5-10B is the exception, answering in free-form prose on
a subset of clips from which no label can be extracted.

\paragraph{TATP.} The TATP prompt supplies the 15 observed frames together with
the corresponding boxes in absolute pixel coordinates, and requests exactly 45
lines in the same format
(Figs.~\ref{fig:prompt_tatp_mtw}--\ref{fig:prompt_tatp_bus}). Three response
types occur. Most models return the requested 45 boxes. Others return a
well-formed but incorrect number of boxes; where more than 45 are produced, the
sequence is truncated to the first 45 and scored normally, which is how the
affected models still appear in Table~3 of the main paper. Answers that cannot be parsed are
scored as the last observed box. One model cannot be scored at all:
Alpamayo-1.5-10B produces no boxes, describing the scene in prose instead, and
is reported as a dash.}

{
\subsection{Human Baseline}
\label{sec:supp_human}

To characterize the difficulty of ERMA independently of any model, we measured human performance on the same task. Five participants who had no prior involvement in the project or the annotation effort, three male and two female,
each answered 20 questions drawn at random from a corpus of 50 clips sampled from the test split. Participants were shown 16 frames sampled within the $\tau=0.5$ observation window and asked to select one of the seven maneuver
classes, with no replay and no additional context. The 50 clips and the subset seen by each participant were balanced across the seven classes.

Human accuracy averaged $79.0\%$ with a macro-F1 of $74.5$, above every model evaluated on the same 50 clips, including the strongest fine-tuned backbone (X3D, $54.0\%$). Individual scores ranged from $60\%$ to $90\%$,
so the mean carries substantial uncertainty at five participants. The results indicate that rider intent is recoverable from a half-observed
maneuver by an untrained observer. The gap to human performance is large for every model family: 25.0 points for the strongest fine-tuned backbone and 47.0 points for the strongest zero-shot model (UniDriveVLA, $32.0\%$).
}

\end{document}